\documentclass[10pt,journal,compsoc]{IEEEtran}

\usepackage[utf8]{inputenc}
\usepackage[T1]{fontenc}
\usepackage{cite}
\usepackage{amsmath,amssymb,amsfonts}
\usepackage{graphicx}
\usepackage[table,dvipsnames]{xcolor}
\usepackage{booktabs}
\usepackage{multirow}
\usepackage{makecell}
\usepackage{array}
\usepackage{adjustbox}
\usepackage{algorithm}
\usepackage{algpseudocode}
\usepackage[normalem]{ulem}
\usepackage{soul}
\usepackage{wrapfig}
\usepackage{float}
\usepackage{caption}
\usepackage{subcaption}
\usepackage{stfloats}
\usepackage{url}
\usepackage{hyperref}

\hypersetup{
    colorlinks=true,
    linkcolor=black,
    citecolor=black,
    urlcolor=black
}

\newcommand{\graycell}{\cellcolor{gray!20}}
\newcolumntype{C}[1]{>{\centering\arraybackslash}p{#1}}
\newcommand{\SAop}{\mathrm{SA}}
\newcommand{\CAop}{\mathrm{CA}}
\newcommand{\SAx}[1]{\SAop\!\left(#1\right)}
\newcommand{\CAx}[2]{\CAop\!\left(#1\,\middle|\,#2\right)}

\newcommand{\qms}[1]{\textcolor{red}{#1}}

\title{Multi-Modal Scene Graph with Kolmogorov--Arnold Experts for Audio-Visual Question Answering}

\author{
        Mengshi Qi,~\IEEEmembership{Member,~IEEE},
        Zijian Fu,
        Changsheng Lv,
        Xianlin Zhang,
        Huadong Ma,~\IEEEmembership{Fellow,~IEEE}
\thanks{This work is partly supported by the Funds for the NSFC Project under Grant 62572072, Beijing Natural Science Foundation (L243027). (\emph{Corresponding author: Mengshi Qi~(email:~qms@bupt.edu.cn)})}
\thanks{M. Qi, Z. Fu, C. Lv, X. Zhang and H. Ma are with the State Key Laboratory of Networking and Switching Technology, Beijing University of Posts and Telecommunications, China.}    

}

\IEEEtitleabstractindextext{%
\begin{abstract}
  In this paper, we propose a novel Multi-Modal Scene Graph with Kolmogorov–Arnold Expert Network for Audio-Visual Question Answering (SHRIKE). The task aims to mimic human reasoning by extracting and fusing information from audio-visual scenes, with the main challenge being the identification of question-relevant cues from the complex audio-visual content. Existing methods fail to capture the structural information within video, and suffer from insufficient fine-grained modeling of multi-modal features. To address these issues, we are the first to introduce a new multi-modal scene graph that explicitly models the objects and their relationship as a visually grounded, structured representation of the audio-visual scene, yielding $461{,}292$ relation triplets over $9{,}288$ musical performance videos. Furthermore, we design a Kolmogorov–Arnold Network~(KAN)-based Mixture of Experts (MoE) to enhance the expressive power of the temporal integration stage. This enables more fine-grained modeling of cross-modal interactions within the question-aware fused audio-visual representation, leading to capture richer and more nuanced patterns and then improve temporal reasoning performance. We evaluate the model on the established MUSIC-AVQA and MUSIC-AVQA v2 benchmarks, where it achieves state-of-the-art performance, reaching $78.14\%$ average accuracy on MUSIC-AVQA, surpassing the previous best method QA-TIGER and ranking first under all four configurations of MUSIC-AVQA v2.0. Code and model checkpoints will be publicly released at \url{https://github.com/feel12348/SHRIKE}.
\end{abstract}
}

\begin{document}

\maketitle

\IEEEdisplaynontitleabstractindextext
\IEEEpeerreviewmaketitle

\section{Introduction}
\label{sec:intro}

Audio-Visual Question Answering (AVQA) aims at answering questions by simulating the human ability to interpret audiovisual scenes by capturing useful auditory and visual cues.
This form of multimodal reasoning remains a significant challenge for machine intelligence, yet it is essential for applications such as robot navigation~\cite{yang2024rila}, embodied intelligence~\cite{ng2024audio}, autonomous driving~\cite{furletov2021auditory} and video surveillance~\cite{gao2024audio}, as well as video retrieval and summarization~\cite{fuentes2022urban}.

\begin{figure}[!t]
    \centering
    \begin{minipage}[b]{0.48\textwidth}
        \centering
        \includegraphics[width=1.0\textwidth]{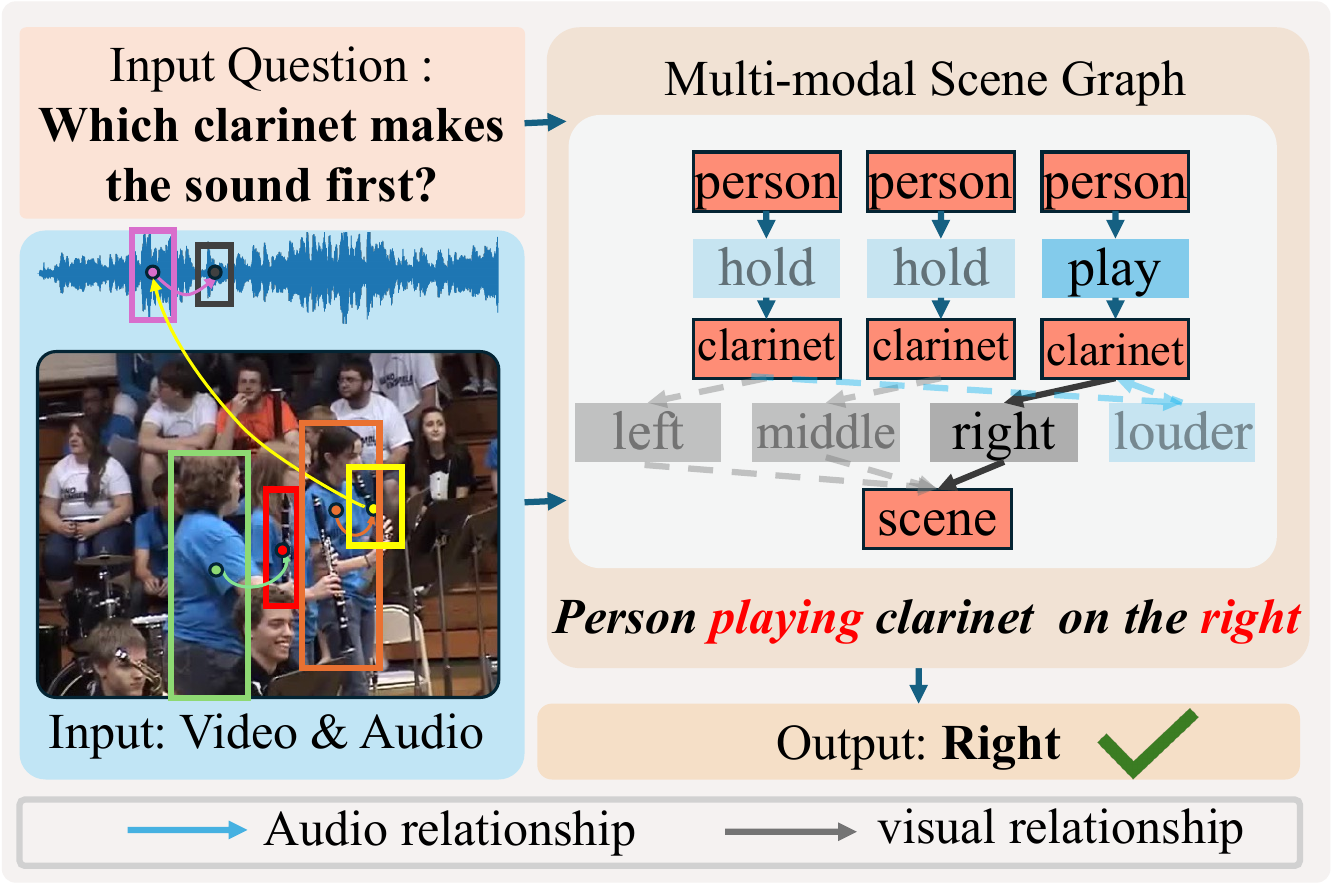}
        \label{teaser4}
    \end{minipage}
    \vspace{-4mm}
    \caption{Illustration of AVQA task. Given an video, we construct a multi-modal scene graph that encodes objects, visual and audio relationships. The textual question is then used to select the most relevant relationships, forming a question-conditioned subgraph that is fed into the fusion and reasoning module to output the accurate answer.}
    \label{teaser}
\end{figure}

Most of existing methods for AVQA were introduced to address challenges like spatio-temporal grounding~\cite{li2022learning,jiang2025clip}, cross-modal adaptation of pre-trained visual backbones~\cite{lin2023vision,duan2023cross,wang2024towards}, and question-guided temporal segment selection~\cite{chen2023question,ma2024look}, yet often overlook visual structural information, particularly when overlapping audio cues make it too hard to distinguish individual sounds. For example, as shown in Figure~\ref{teaser}, answering the question ``Which clarinet makes the sound first?'' based solely on audio can be ambiguous due to sound blending. In contrast, learning from multimodal structural relationships, such as spatial arrangement, can provide clearer cues for accurate identification. Therefore, we are the first to extend the current scene graph~\cite{johnson2015image} into the multi-modal field~(\emph{i.e.,} image, text and audio) to explicitly model audio-visual structural relationships, incorporating structural context as complementary information to enhance overall audio-visual understanding.

Furthermore, another key challenge lies in the temporal grounding of question-relevant cues. Existing methods~\cite{2025tiger} adopt a Mixture-of-Experts (MoE)~\cite{shazeer2017outrageously} framework with MLP-based experts to capture temporal dependencies through adaptive weighting. While such MLP-based experts are effective at modeling coarse global temporal trends, the learned temporal weights remain implicit and predominantly coarse-grained, making them insufficient for aggregating fine-grained, question-relevant cross-modal cues. However, the human has the intrinsic capacity to jointly reason over multimodal details and global contexts, the recently introduced Kolmogorov-Arnold Networks (KANs)~\cite{kolmogorov1957elements}, with its adaptive edge‑wise nonlinearities and strong data-fitting capabilities, offers a viable means to emulate this distinctive aspect of human inference. Compared to MLPs, KANs exhibit stronger locality under the same network depth, owing to their spline-based parameterization. This locality allows the temporal integration module to more easily focus on question-relevant events, enabling finer-grained modeling of cross-modal temporal patterns and ultimately improving temporal reasoning performance. To the best of our knowledge, we are the first to explore KAN-based experts for audio–visual question answering.

In this paper, we propose a new Multi-Modal \textbf{S}cene G\textbf{R}ap\textbf{H} w\textbf{I}th \textbf{K}olmogorov–Arnold framework for Audio-Visual Qu\textbf{E}stion Answering, termed as \textbf{\emph{SHRIKE}}. First, features are extracted from the input video, audio, and questions using their respective encoders. Next, we introduce a new \textbf{\emph{Multi-Modal Scene Graph Decoder~(M$^2$SG)}}, which generates triplet features representing structural information in each frame based on its audio and visual features. These features are further fused with the temporally grounded visual and audio features via the cross-attention mechanism, serving as enhanced audio cues and complementary structural context. Finally, a classifier is employed to predict the answer to the question. Meanwhile, within the MoE framework, we introduce a new \textbf{\emph{Kolmogorov-Arnold Network (KAN) based Expert Module}} to enhance the expressive power of temporal integration. Specifically, KAN is employed to replace MLPs within the feed-forward networks (FFNs) that constitute the diverse experts in the MoE. By concentrating multiple Gaussian distributions along the temporal axis, KAN learns and adaptively aggregates the weights of these distributions, generating a soft mask that facilitates the extraction of sharper and more reliable temporal dependencies. These two newly-designed components are complementary within the proposed framework: the multi-modal scene graph specifies what relational structure should be reasoned over, while the KAN-based expert module determines how temporally relevant evidence should be modulated and aggregated, thereby yielding a unified and task-oriented solution for AVQA.

Our main contributions can be summarized as follows:

\textbf{(1)} We propose a novel Multi-Modal Scene Graph framework~({\it i.e.}, SHRIKE), integrated with the Kolmogorov–Arnold Network (KAN) for Audio-Visual Question Answering, which enhances reasoning capabilities through the newly designed Multi-Modal Scene Graph Decoder, particularly in scenarios with overlapping and hard-to-distinguish audio cues.

\textbf{(2)} We introduce a new KAN-based expert module in the MoE framework, replacing MLPs with spline-based KAN experts to enhance the expressive power of the temporal integration and reasoning performance.

\textbf{(3)} We construct the first large-scale multi-modal scene graph dataset based on MUSIC-AVQA through a cascaded three-stage annotation pipeline with rule-based verification and regeneration, providing $461{,}292$ relation triplets over $92{,}880$ annotated segments of $9{,}288$ videos as segment-level structural supervision for audio-visual reasoning.

\textbf{(4)} Experimental results show our proposed SHRIKE achieves state-of-the-art performance on the MUSIC-AVQA and MUSIC-AVQA 2.0 datasets, reaching $78.14\%$ average accuracy on MUSIC-AVQA, while providing explicit temporal grounding of audio and visual cues.


\section{Related Work}
\label{sec:formatting}

\textbf{Audio-Visual Scene Understanding.} Inspired by human multisensory perception, research has increasingly focused on audio-visual scene understanding, spanning tasks such as action recognition~\cite{kazakos2019epic,gao2020listen}, audio-visual event localization~\cite{tian2018audio}, sound source segmentation~\cite{zhou2022audio,qi2026dcsam}, question answering~\cite{alamri2019audio}, and object localization~\cite{huang2023egocentric}. Among these, audio-visual question answering (AVQA) is particularly challenging, as it requires jointly perceiving heterogeneous modalities and reasoning over their interactions~\cite{qi2026explainable,qi2025action,qi2026balanced} to answer free-form natural-language questions. Early attempts adapt video question answering pipelines by fusing modality-specific features with recurrent or attention-based encoders~\cite{fayek2020temporal,zhou2016attention,lu2016hierarchical,yu2019deep}, while ST-AVQA~\cite{li2022learning} establishes a large-scale spatio-temporal grounding benchmark on musical performances and provides a strong baseline. Building on this, subsequent works improve AVQA from complementary perspectives. LAVISH~\cite{lin2023vision} introduces adapter-based architectures that generalize pre-trained visual backbones to audio-visual scenarios with minimal additional parameters, whereas COCA~\cite{lao2023coca} and M2KVDG~\cite{liu2024m2k} exploit causal graphs~\cite{qi2026rdcl} to suppress spurious correlations and improve multi-modal interaction and reasoning. Another line of research targets question-guided temporal grounding, including progressive spatio-temporal perception~\cite{li2023progressive}, key semantic-aware sampling~\cite{li2024boosting}, object-aware adaptive positivity learning~\cite{li2024object}, and question-conditioned temporal segment selection~\cite{chen2023question,ma2024look}; others further explore cross-modal adaptation of large pre-trained backbones~\cite{duan2023cross,wang2024towards,jiang2025clip}. Despite their progress, most of these methods treat the audio-visual scene as unstructured feature sequences and overlook the explicit relational structure among sounding objects, which is precisely what makes reasoning ambiguous when multiple instruments sound simultaneously. In contrast, our method integrates graph-structured audio-visual information that explicitly encodes objects and their cross-modal relations, providing complementary structural cues that are especially valuable when overlapping sounds are hard to disambiguate.

\noindent\textbf{Scene Graph~(SG).} SGs were originally introduced for image retrieval~\cite{johnson2015image} as a structured representation that decomposes a scene into objects and their pairwise relations, and were later scaled up by large-scale relational annotations such as Visual Genome~\cite{krishna2017visual}. By explicitly modeling entity interactions, SGs provide richer and more interpretable semantic context than global feature embeddings, and have been widely adopted in image retrieval~\cite{yoon2021image}, visual question answering~\cite{qian2022scene}, and captioning~\cite{li2019know}. Recent works further demonstrate their effectiveness in more complex settings~\cite{lv2025t2sg}, such as 3D scene retrieval~\cite{chen2024scene} and structured compositional reasoning on CLEVR~\cite{shi2019explainable,johnson2017clevr}. From a modeling perspective, recent scene graph generators~\cite{lv2026robosgg} increasingly cast relation prediction as a set-prediction problem using query-based decoders inspired by DETR~\cite{carion2020end}, which directly predict a set of relation triplets via learnable queries and bipartite matching, avoiding hand-crafted proposal heuristics. In this work, our proposed Multi-Modal Scene Graph Decoder follows this query-based paradigm, but extends it to jointly reason over visual and audio evidence. Crucially, existing SG-based methods are confined to purely visual scenarios and cannot express auditory relations, such as the relative loudness between instruments or whether an instrument is actually being played rather than merely held. To the best of our knowledge, we are the first to extend scene graphs to the multi-modal (visual--audio) setting and inject them into AVQA, enabling more robust reasoning under ambiguous and overlapping audio cues.

\noindent\textbf{Mixture of Experts~(MoE).}
MoE~\cite{shazeer2017outrageously} enhance model capacity without a proportional increase in computation by routing each input to the specialized expert sub-networks through a learnable gating function. Owing to this conditional-computation property, MoE has become a key ingredient for scaling large models efficiently, as demonstrated by GShard~\cite{lepikhin2020gshard} and GLaM~\cite{du2022glam}. Beyond efficiency, the inherent specialization of different experts makes MoE attractive for modeling the heterogeneous and multi-faceted patterns that arise in multi-modal data. In AVQA, QA-TIGER~\cite{2025tiger} instantiates the experts as MLP-based feed-forward networks and couples them with Gaussian temporal weighting to softly aggregate question-relevant temporal segments. However, MLP experts tend to capture coarse and globally smooth temporal trends, which limits their ability to localize fine-grained, question-specific cross-modal cues. Motivated by this limitation, we retain MoE routing formulation but replace MLP experts with spline-parameterized Kolmogorov--Arnold experts, whose stronger locality yields sharper and more reliable temporal aggregation.

\noindent\textbf{Kolmogorov-Arnold Networks~(KANs).} The Kolmogorov--Arnold theorem~\cite{kolmogorov1957elements} states that any multivariate continuous function can be represented as a finite composition of univariate functions and additions, which has recently inspired a new family of neural architectures known as KANs~\cite{liu2024kan}. Unlike MLPs that apply fixed nonlinearities on nodes and learn linear weights on edges, KANs place learnable spline-based activation functions on edges, leading to stronger local expressiveness, fewer parameters, and improved interpretability~\cite{liu2024kan}. Recent extensions broaden their applicability: TimeKAN performs frequency decomposition for time-series modeling~\cite{huang2025timekan}, while PowerMLP matches the efficiency of MLPs while retaining the expressiveness of KANs~\cite{qiu2025powermlp}. The spline-based parameterization endows KANs with a locality property that makes them naturally suited to concentrate on a few question-relevant temporal intervals rather than smoothing over the entire sequence. To the best of our knowledge, we are the first to leverage KAN-based experts to enhance the expressive power of temporal integration for modeling complex cross-modal temporal dependencies in AVQA.

\section{Multi-Modal Scene Graph Dataset}
\label{Sec: MMSG Dataset}

We construct the first large-scale multi-modal scene graph dataset on
top of MUSIC-AVQA~\cite{li2022learning}. The dataset provides
question-independent, segment-level structural supervision for
pre-training the Multi-Modal Scene Graph Decoder in
Section~\ref{Sec: Multi-Modal Scene Graph Generation}. This section
first defines the dataset task, then summarizes the construction
pipeline, and finally reports the statistics of the resulting dataset.
Detailed data preparation, ontology design, annotation prompts, and
verification rules are provided in the supplementary material.

\subsection{Problem Setting}
\label{Sec: Problem Setting}

We first distinguish the dataset annotation task from the downstream
audio--visual question answering (AVQA) task. In AVQA, a model receives
a video $\mathcal{V}$, its aligned audio $\mathcal{A}$, and a question
$q$, and predicts an answer $y$ from a candidate set $\mathcal{C}$:
$y\in\mathcal{C}$. While our dataset provides
question-independent structural annotations. The question is not used
when constructing a scene graph, so the same annotation can support
multiple questions associated with the video. Let $\mathcal{O}$ and $\mathcal{R}$ denote the entity and predicate
vocabularies, respectively. We use a compact, task-oriented ontology
whose entity and predicate categories, together with their type
constraints.

We uniformly divide each audio--visual stream into $T$
non-overlapping temporal segments. For segment $t$, we define a
multi-modal scene graph as
$g_t=(\mathcal{N}_t,\mathcal{E}_t)$, where
$\mathcal{N}_t$ is the set of entity instances appearing in the
segment and $\mathcal{E}_t$ is the set of typed relation edges. Each
edge is represented as a subject--predicate--object triplet:
\begin{equation}
\mathcal{E}_t=
\left\{e_{t,i}=\langle s_{t,i},p_{t,i},o_{t,i}\rangle
\right\}_{i=1}^{K_t},
\end{equation}
where $s_{t,i},o_{t,i}\in\mathcal{N}_t$,
$p_{t,i}\in\mathcal{R}$, and $K_t$ is the number of valid triplets in
the segment. The graph sequence for the complete video is
$\mathcal{G}=\{g_t\}_{t=1}^{T}$. Consequently, the constructed dataset
can be written as the following:
\begin{equation}
\mathcal{D}_{\mathrm{MMSG}}=
\left\{(\mathcal{V}^{(n)},\mathcal{A}^{(n)},
\mathcal{G}^{(n)})\right\}_{n=1}^{9{,}288}.
\end{equation}

Each triplet follows the type constraints of the ontology, whose full
definition is deferred to the supplementary material. A special
\textit{nothing} label is assigned when a segment contains no
admissible triplet. These annotations supervise a
question-independent mapping from each audio--visual segment
$(\mathcal{V}_t,\mathcal{A}_t)$ to its triplet set $\mathcal{E}_t$.
The predicted graph sequence can then be combined with a question by
the downstream AVQA model, as described in
Section~\ref{Sec: Multi-Modal Scene Graph Generation}.

\subsection{Dataset Construction}
\label{Sec: Dataset Construction}

Our dataset covers all $9{,}288$ one-minute videos in
MUSIC-AVQA~\cite{li2022learning}. Each video is normalized to a
$60$-second timeline and divided into ten non-overlapping $6$-second
segments, yielding $92{,}880$ annotation units. Construction follows a
cascaded three-stage pipeline. Stage~1 builds a video-level entity
inventory with persistent instance identifiers. Stage~2 conditions
MiniCPM-o~2.6~\cite{yao2024minicpm} on this inventory to generate an
initial graph from the key frame and synchronized audio of each
segment. Stage~3 uses Qwen3-Omni-30B-A3B
(Thinking)~\cite{xu2025qwen3} to jointly refine all ten graphs from the
complete audio--visual sequence, improving temporal consistency and
entity re-identification across segments.

The generated outputs are subsequently processed by an automatic
verification-and-regeneration loop. Structural validators reject
truncated, empty, or malformed outputs; an ontology filter normalizes
synonyms and removes triplets that violate the typed relation forms;
and an LLM-based relevance check compares the graphs against the
available question--answer evidence. Invalid annotations are
regenerated, while human experts inspect and correct a sampled subset.
The resulting per-video sequence of ten triplet sets is used as ground
truth for pre-training the Multi-Modal Scene Graph Decoder. The full
preparation and annotation procedures are documented in the
supplementary material.

\textbf{Cost and Quality of Annotations Generated by
Foundation Models.} The scene graph annotations are generated entirely
offline using 8 RTX 4090 GPUs, at an average cost of about 3 minutes
per video. It is worth emphasizing that this cost is incurred only once
during the construction of the training annotations; since no MLLM is
involved during inference, the runtime overhead of our method remains
comparable to that of QA-TIGER~\cite{2025tiger}, and detailed
comparisons are given in the supplementary material. Furthermore, the
raw scene graphs are not used directly, but are further refined by an
automated correction pipeline that combines rule-based constraints,
LLM-based relevance validation, and expert correction on a subset of
samples.

\subsection{Dataset Statistics}
\label{Sec: Dataset Statistics}

\begin{figure*}[t]
\centering
\includegraphics[width=0.98\textwidth]{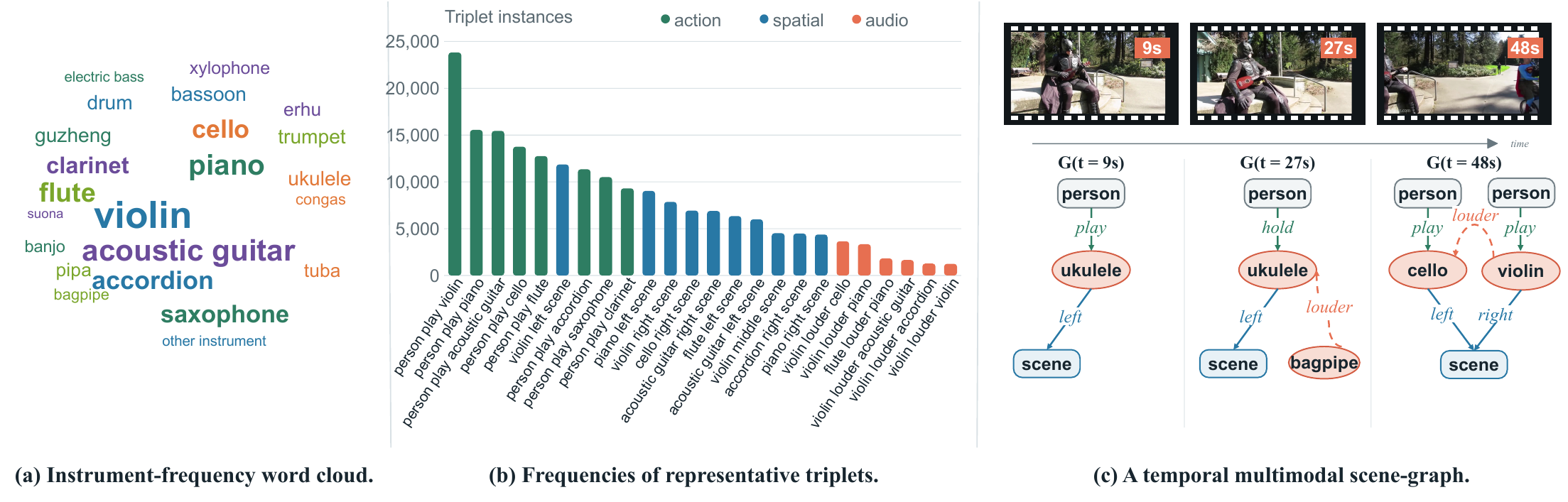}
\caption{Overview of our multi-modal scene graph dataset. (a) An
instrument-frequency word cloud, where font size indicates the number
of occurrences. (b) Frequencies of representative triplets, grouped
by relation modality. (c) A temporal example showing synchronized
video observations and their evolving multi-modal scene graphs at
three timestamps.}
\label{Fig: dataset_overview}
\end{figure*}

The final dataset contains $9{,}288$ videos and $92{,}880$ annotated
$6$-second segments with $461{,}292$ relation triplets in total,
corresponding to an
average of $4.97$ triplets per segment. In total, $560$ of the $644$
admissible triplet categories are observed, while only $1{,}896$
segments ($2.0\%$) contain no admissible triplet. These statistics
indicate that the annotations provide dense and diverse structural
supervision. Detailed category definitions and distributions are
reported in the supplementary material.

Figure~\ref{Fig: dataset_overview} provides a complementary view of
the dataset composition. As shown in Figure~\ref{Fig: dataset_overview}(a),
the entity distribution is long-tailed: violin and acoustic guitar are
the most prominent instruments, followed by accordion, piano, and
several other common instrument categories. Figure~\ref{Fig: dataset_overview}(b)
further shows that action triplets dominate the high-frequency portion
of the distribution, while spatial and auditory triplets provide
additional structural coverage. Finally,
Figure~\ref{Fig: dataset_overview}(c) illustrates how the annotations
evolve with the audio--visual content over time, transitioning from a
single-instrument scene to interactions involving multiple instruments.
It highlights that our annotations capture not only static
entities but also temporally varying cross-modal relations.

\section{Method}
\label{headings}

\begin{figure*}[htbp]
  \centering
  \includegraphics[width=0.95\linewidth]{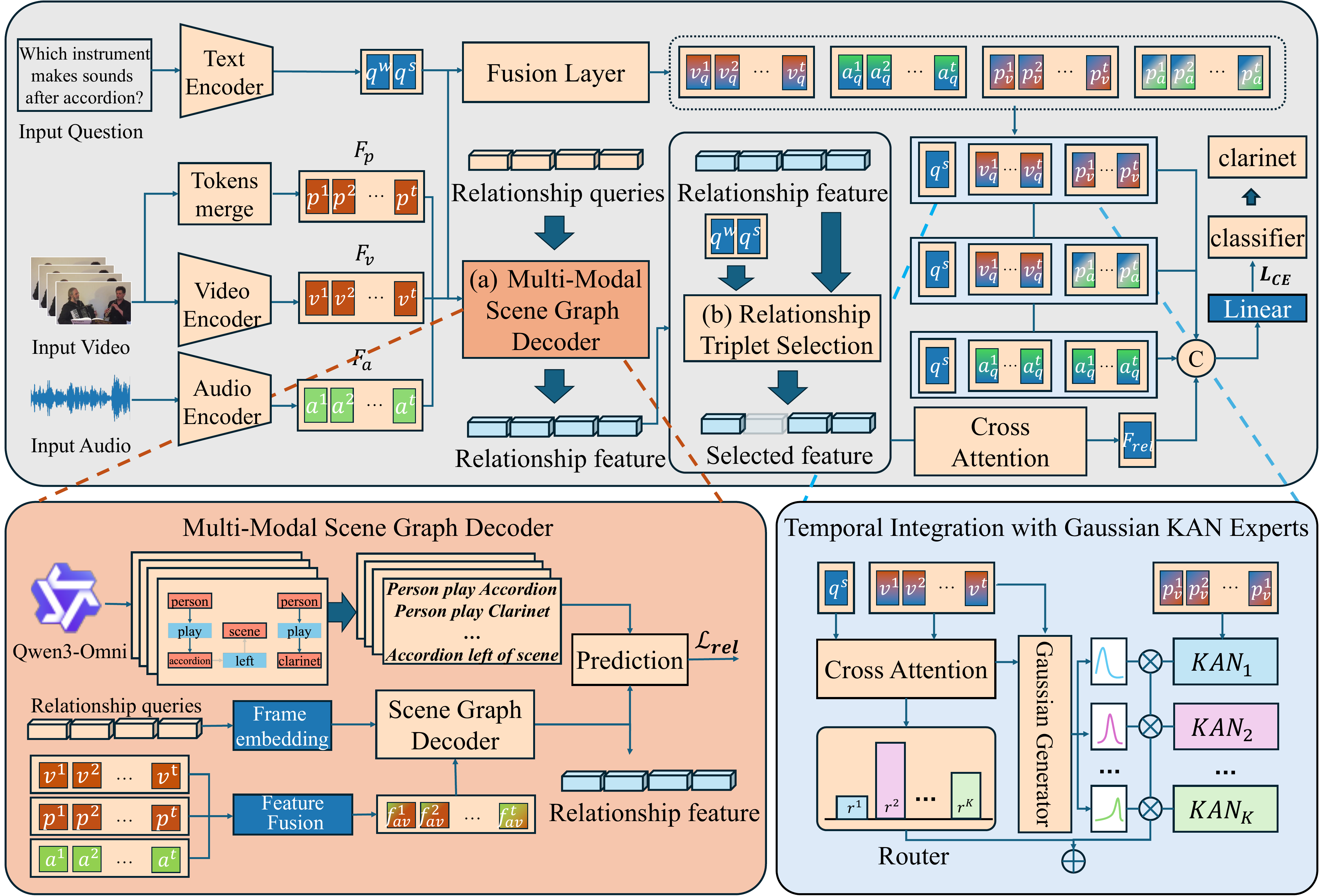}
  \caption{Overview of our proposed SHRIKE: Features from each modality are obtained by passing the input through a corresponding pretrained encoder. Then we propose (a) Multi-Modal Scene Graph Decoder to extract scene graph features from the video and select specific triplets using the relationship triplets selection. Through (b) Temporal Integration with Gaussian KAN Experts, our model achieves effective question-guided localization of critical temporal segments, enhancing temporal reasoning and multi-modal understanding.}
  \label{Fig: Main Model}
\end{figure*}

\subsection{Overview}
\label{Sec: overview}
As shown in Figure~\ref{Fig: Main Model}, our proposed SHRIKE is designed to advance audio–visual question answering by (i) constructing an explicit multi-modal scene graph that provides structured entity–relation cues, and (ii) introducing a KAN-based expert module that enhances temporal integration of audio and visual evidence. Our model is optimized in a two-stage manner. In the first stage, Multi-Modal Scene Graph Decoder is trained; while in the second stage, this decoder is frozen, and the remaining components of the model are trained conditioned on the predicted scene graphs.

\noindent\textbf{Problem Formulation.} Given an audio--visual question answering input $\langle\mathcal{V},\mathcal{A},q_s\rangle$, where $\mathcal{V}$, $\mathcal{A}$, and $q_s$ denote the video, audio, and question, respectively, our model predicts an answer $y\in\mathcal{C}$ from a predefined candidate set $\mathcal{C}$. This downstream AVQA formulation is distinct from the question-independent scene graph annotation task defined in Section~\ref{Sec: Problem Setting}.

\noindent\textbf{Scene Graph Supervision.} We adopt the notation and ontology defined in Section~\ref{Sec: Problem Setting}. For each temporal segment $t$, the Multi-Modal Scene Graph Decoder learns to predict a triplet set $\widehat{\mathcal{E}}_t$ from the corresponding audio--visual input $(\mathcal{V}_t,\mathcal{A}_t)$ without observing the question. The predicted segment-level graphs form $\widehat{\mathcal{G}}=\{\widehat{g}_t\}_{t=1}^{T}$ and provide question-independent structural cues to the subsequent AVQA modules. The full construction procedure of these annotations is detailed in Section~\ref{Sec: MMSG Dataset}, and the detailed annotation procedures are provided in the supplementary.

\noindent\textbf{Input Representations.} Given a video $\mathcal{V}$ and its corresponding audio $\mathcal{A}$, we follow~\cite{2025tiger} and employ pre-trained audio, visual, and text encoders with frozen parameters to extract features. For each segment $t$: \textbf{(1) Audio.} We use the pre-trained VGGish model~\cite{vgg2017} to obtain audio features $a_t \in \mathbb{R}^{D}$, forming 
$F_a = \{a_t\}_{t=1}^{T} \in \mathbb{R}^{T \times D}$.
\textbf{(2) Visual.} A pre-trained CLIP visual encoder~\cite{clip2021} is used to extract frame-level features 
$F_v = \{v_t\}_{t=1}^{T} \in \mathbb{R}^{T \times D}$ from the [CLS] token, and patch-level features 
$F_p = \{p_t\}_{t=1}^{T} \in \mathbb{R}^{T \times M \times D}$ via Token Merging~\cite{2022tome}, where $M$ is the number of visual tokens.
\textbf{(3) Text.} The input question is tokenized and passed through the CLIP text encoder to obtain a sentence-level feature 
$q_s \in \mathbb{R}^{D}$ from the [EOT] token and word-level features 
$q_w \in \mathbb{R}^{M' \times D}$, where $M'$ denotes the token number.
\label{Sec: Multi-Modal Scene Graph}

\subsection{Multi-Modal Scene Graph Generation}
\label{Sec: Multi-Modal Scene Graph Generation}

In this section, we introduce the fusion of audio and visual features, along with the newly designed Scene Graph Decoder for triplet extraction in the scene graph, and describe the process of relationship triplet selection.\\
\noindent\textbf{Feature Fusion.} To comprehensively capture all potential relationships within the scene graph, we intentionally exclude question-related textual features from the model input and instead focus solely on fusing visual and auditory modalities. Incorporating textual information may introduce bias or unintended priors that could interfere with accurate scene graph prediction. Specifically,
we employ cross-attention (CA) and self-attention (SA) mechanisms to fuse visual features $F_v$ and audio features $F_a$ as follows:
\begin{align}
F'_v &= F_v + \SAx{F_v} + \CAx{F_v}{F_a},\\
F'_a &= F_a + \SAx{F_a} + \CAx{F_a}{F_v},
\end{align}

\noindent where $\mathrm{CA}$ and $\mathrm{SA}$ denote cross-attention and self-attention, respectively. The tensors $F'_v, F'_a \in \mathbb{R}^{T \times D}$ represent the enhanced visual and audio features.  To achieve finer spatial granularity and detailed object property identification, we concatenate \( {F'_v} \) and \( {F'_a} \), and apply cross-attention using the concatenated feature as the query and patch-level visual features ${F_p}$ as the key and value, formulated as:
\begin{align}
{F_{av}} &= \CAx{Concat({F'_v},{F'_a})}{F_p},
\end{align}

\noindent where ${F_{av}} \in \mathbb{R}^{T \times D}$ represents the fused feature, and $\mathrm{Concat}$ denotes the concatenation operation.

\noindent\textbf{Scene Graph Decoder.} We initialize learnable relationship queries ${Q} \in \mathbb{R}^{T \times L \times D}$, where $T$, $L$, and $D$ denote the number of frames, queries, and hidden dimension size, respectively. Inspired by Frame Embedding~\cite{urooj2023learning}, we augment the queries with positional encodings to encode temporal information. These queries serve as the input to a three-layer Transformer decoder, where the fused audiovisual features ${F_{av}}$ are used as keys and values. For the $l$-th layer:  
\begin{gather}  
{Q}^{(l+1)} = \mathrm{TransformerDecoderLayer}({Q}^{(l)},\,{F_{av}}),
\end{gather}  
Then we can achieve the final layer output, denoted as ${F_{rel}} = {Q}^{final} \in \mathbb{R}^{T \times L \times D}$, representing triplet relationship features in the audiovisual scene. During training, these features are then passed through a prediction head implemented as a two-layer feed-forward network (FFN) with a GELU activation function.

\noindent\textbf{Relationship Triplet Selection.} The real-world scene contains multiple relationship triplets, but not all of them are relevant to the given question. To address this, we introduce a Relationship Triplet Selection strategy to identify the most relevant triplets. Specifically, for each time segment $t$, we compute the attention weights between the question feature ${q_s}$ and the relationship triplet features ${F_{rel}^t}$ using scaled dot-product attention. The segments with the highest attention weights, along with their corresponding indices, are then selected as the following:  
\begin{gather}  
    {W_{rel}} = \text{softmax}\left( \frac{{q_s} {F_{rel}^t}^\top}{\sqrt{D}} \right), \\ {F^t_{top_{k}}} ={{\phi}_{QTP}}({F^t_{rel}},{W_{rel}},{top_{k}}),
\end{gather}  

\noindent where ${W_{rel}} \in \mathbb{R}^{1 \times N}$ represents the attention weights, and $D$ is the dimension of ${q_s}$. The selection operation is denoted as ${{\phi}_{QTP}}$, and the resulting top-$k$ features are represented as ${F^t_{top_{k}}} \in \mathbb{R}^{top_{k} \times D}$.

\subsection{Fusion Layer}
\label{Sec: Fusion layer}

Subsequently, we integrate information from the video, audio, and text modalities. Specifically, we use self-attention to enhance intra-modal relationships, and then apply cross-attention with the other two modalities to achieve inter-modal fusion: visual features ${F_v}$ are used as queries, while audio features ${F_a}$ serve as keys and values. We employ word-level question features ${q_w}$ as keys and values. Audio features ${F_a}$ are processed in the similar manner. Finally, the resulting features are combined with the original input via a residual connection, as follows:
\begin{align}
v_q &= F_v + \SAx{F_v} + \CAx{F_v}{F_a} + \CAx{F_v}{q_w},\\
a_q &= F_a + \SAx{F_a} + \CAx{F_a}{F_v} + \CAx{F_a}{q_w},
\end{align}

\noindent where ${v_q} = \{{v}_q^{t}\}_{t=1}^{T} \in \mathbb{R}^{T \times D}$ and ${a_q} = \{{a}_q^{t}\}_{t=1}^{T} \in \mathbb{R}^{T \times D}$. 

Afterwards, the patch-level visual features ${F_p}$ is first refined by self-attention with a residual connection to capture the contextual dependencies among patch tokens. Then the fused segment-level features ${v_q}$ and ${a_q}$ act as queries in cross-attention over the refined patch tokens ${F'_p}$ (serving as keys and values), so that each modality aggregates the question-relevant, spatially fine-grained patch information. The process can be described as follows:
\begin{gather}
{F'_p} =  {F_p} + \mathrm{SA}({F_p}), \\
{p_v} =\mathrm{CA}({v_q} ~|~{F'_p}), ~{p_a} =\mathrm{CA}({a_q} ~|~{F'_p}),
\label{eql:3}
\end{gather}

\noindent where ${p_v} = \{{p}_v^{t}\}_{t=1}^{T} \in \mathbb{R}^{T \times D}$ and ${p_a} = \{{p}_a^{t}\}_{t=1}^{T} \in \mathbb{R}^{T \times D}$. Thus we obtain the audio and visual modalities enriched with contextual information from the question. 

\subsection{Gaussian KAN-based Experts}  
\label{Sec: Temporal Integration of Gaussian KAN Experts}  
Inspired by LLMs~\cite{lepikhin2020gshard,du2022glam}, we adopt the Mixture of Experts (MoE)~\cite{shazeer2017outrageously} framework in our model, by leveraging a Gaussian distribution to model temporal patterns. To enhance the expressive power of temporal integration, we are first to leverage the Kolmogorov-Arnold Network (KAN) as the expert network in our model.

\noindent\textbf{Gaussian Generator.} In order to distribute the Gaussian centers of multiple expert models across the temporal dimension, we adopt the cross-attention to align the question features ${q_s}$ with the fused multimodal features ${v_q}$ in Section~\ref{Sec: Fusion layer}. The resulting features $v'_{q}$ are then projected through a fully connected layer (\text{FC}) to map the input dimension $D$ to a 2-dimensional output, yielding the mean ($\mu$) and standard deviation ($\sigma$) of the Gaussian distribution, formalized as follows:  
\begin{align}  
    {v'_{q}} = \CAx{q_s}{v_q}, \, \mu_{v}, \sigma_{v} = \text{FC}({v'_{q}}), \, {\delta} = \mathcal{N}(\mu_{v}, \sigma_{v}^2).
\end{align} 

\noindent\textbf{Integrating Temporal Information.} 
We then employ Kolmogorov -Arnold Networks~(KAN) to replace MLPs within each expert, aiming to enhance the expressive power of temporal integration. Let \(\mathcal{F}_{\text{KAN}}({x})\) denote outputs of KAN transformation applied to the input \(d\)-dimensional features \({x} = [x_1, x_2, \ldots, x_d]\), formulated as: 
\begin{gather}  
    \mathcal{F}_{\text{KAN}}({x_{i}}) = \sum_{i=1}^{d} \left( \sum_{k=1}^{K} c_{ijk} \, B_k(x_i) \right),  
\end{gather}  
where \(x_i\) represents the \(i\)-th dimension, \(B_k(x_i)\) denotes the \(k\)-th basis function ({\it i.e.}, B-spline) applied to \(x_i\), and \(c_{ijk}\) is the learnable coefficient in the KAN. Leveraging spline-based parameterization offers two main advantages:
1) Owing to the compact support of B-spline basis functions, each coefficient only affects the response within a limited interval of the input space. 
Such a locality property allows the model to adjust nonlinear transformations in a feature-specific and value-dependent manner, which is beneficial for capturing subtle variations in the input representation. 
2) Since splines are piecewise polynomials, the resulting KAN mapping exhibits a favorable inductive bias: it preserves overall smoothness while retaining sufficient local flexibility to model complex nonlinear patterns. 
Compared with standard MLP-based transformations, this property can provide more precise feature discrimination, which is particularly useful for fine-grained event understanding.

To capture temporally critical segments, we adopt a Mixture of Experts (MoE) framework~\cite{2025tiger}. For each KAN-based expert, we design multiple Gaussian distributions \(\{\delta^i\}_{i=1}^{E}\) using Gaussian Generators, where \(E\) denotes the number of experts. For features \(o \in \{a, p_a, p_v\}\), where $p_a$ and $p_v$ are from Equation~(\ref{eql:3}), the procedure is formulated as follows:  
\begin{gather}  
    \widetilde{F_{o}} = \sum_{i=1}^{E} {\delta}_{o}^{i} \, {r_{o}^{i}} \, \mathcal{F}_{\text{KAN}}({x_{o}^{i}}), \,\text{where}\, r_o = \text{FC}(v'_q),
\end{gather}  
Here, \(\{{\delta}_{o}^{i}\}_{i=1}^{E} \in \delta_o\) represents the Gaussian distributions corresponding to feature \(o\), and \(\{r_{o}^{i}\}_{i=1}^E\) denotes the routing scores for feature \(o\), obtained via a fully connected layer mapping from feature dimension \(D\) to the number of experts \(E\). Finally, we obtain the MoE-aggregated features, \emph{i.e.,} \(\widetilde{F_{a}}\), \(\widetilde{F_{p_a}}\), and \(\widetilde{F_{p_v}}\).

\subsection{Optimization}
\label{Optimization}
Finally, our proposed SHRIKE framework incorporates two loss functions: {\bf (1) Task-specific Classification Loss} \(\mathcal{L}_{task}\) for the Audio-Visual Question Answering task. We concatenate the MoE-aggregated features \(\widetilde{F_{a}}\), \(\widetilde{F_{p_a}}\), and \(\widetilde{F_{p_v}}\) with relationship features along the feature dimension and pass them through a fully connected layer to construct a classifier, yielding the predicted answer \(\hat{y}_c\). {\bf (2) Relationship Prediction Loss} \(\mathcal{L}_{rel}\) from the Scene Graph Decoder in Section~\ref{Sec: Multi-Modal Scene Graph Generation}. We employ a Hungarian loss $\mathcal{L}_{\text{match}}$~\cite{carion2020end} to optimally match the predicted relationship triplet \(\hat{r}_{t, i}\) with the ground-truth labels \(r_{t ,i}\) from the Multi-Modal Scene Graph Annotations generated by qwen3-omni~\cite{xu2025qwen3}. This can be expressed as:  
\begin{gather}  
    \hat{\sigma}_{rel} = \arg\min_{\sigma \in \mathfrak{S}_N} \sum_{t=1}^{T}\sum_{i=1}^{N} \mathcal{L}_{\text{match}} \left( y_i^t, \hat{y}_{\sigma(i)}^t \right),\\
    \mathcal{L}_{task} = - \sum_{c=1}^{C} \hat{y_c} \log p_c, \mathcal{L}_{rel} = \sum_{t=1}^{T} \sum_{i=1}^{N} -y_{t,i}\log p_{\hat{\sigma}_{rel(t,i)}}.
\end{gather}  
Specifically, in the first stage, we train the Multi-Modal Scene Graph Decoder using the loss function $\mathcal{L}_{rel}$; in the second stage, we apply the loss function $\mathcal{L}_{task}$ to perform end-to-end training of the entire model.

\section{Experiments}
\subsection{Datasets and Evaluation Metrics}
\noindent\textbf{Dataset.} We evaluate our method on two widely used audio--visual question answering benchmarks, \emph{i.e.,} \emph{MUSIC-AVQA}~\cite{li2022learning} and \emph{MUSIC-AVQA-v2.0}~\cite{liu2024tackling}, both of which are built upon real-world musical performance videos and are specifically designed to test spatio-temporal reasoning across the auditory and visual modalities. \emph{MUSIC-AVQA}~\cite{li2022learning} is a large-scale dataset that contains 9,288 musical performance videos, spanning 22 different instruments and amounting to roughly 150 hours of footage, together with 45,867 manually annotated question--answer pairs. Since these videos frequently involve several instruments being played at the same time, correctly answering the questions often requires jointly perceiving and associating cues from both modalities rather than relying on either one alone, which makes the benchmark a suitable testbed for evaluating fine-grained multi-modal reasoning. The questions are organized into five representative types, namely existential, counting, location, comparative, and temporal questions, and are further grouped, according to the modalities they rely on, into \emph{Audio QA}, \emph{Visual QA}, and \emph{Audio-Visual QA}. We follow the official setting, which partition the question-answer pairs into training, validation, and test sets in a ratio of approximately $7{:}1{:}2$, to ensure a fair comparison with prior work. \emph{MUSIC-AVQA-v2.0}~\cite{liu2024tackling} further extends this benchmark to 10,492 videos and, more importantly, rebalances the answer distribution in order to mitigate the data bias present in the original dataset. It provides both a biased and a balanced subset and supports evaluation under different combinations of training and test distributions. This design enables a more reliable assessment of model robustness under diverse multi-instrument scenes, and helps verify whether the observed gains stem from genuine cross-modal reasoning rather than from exploiting dataset bias.

\begin{table*}[t]
\footnotesize
\centering
\caption{Comparison with state-of-the-art methods on the MUSIC-AVQA~\cite{li2022learning} test set. The best-performing results are shown in \textbf{bold}, and the second-best results are \ul{underlined}. $\dagger$ denotes results obtained by reproducing the model using official codes. $\ddagger$ denotes VideoLLaMA-2 results reported by MAVEN~\cite{ma2025fortisavqa}. $\S$ denotes methods implemented with the same encoder backbone for fair comparison. Dashes indicate results not reported by the corresponding work.
We also conduct additional experiments with other encoder backbones, and report in Appendix.}
\label{tab: music-avqa}
\begin{adjustbox}{width=1.01\linewidth,center=\linewidth}
{
    \fontsize{8}{11}\selectfont
    \begin{tabular}{l|ccc|ccc|cccccc|c}
    \toprule
    \multirow{2}{*}{\textbf{Method}} & \multicolumn{3}{c|}{\textbf{Audio QA}} & \multicolumn{3}{c|}{\textbf{Visual QA}} & \multicolumn{6}{c|}{\textbf{Audio-Visual QA}} & \multirow{2}{*}{\textbf{Avg}} \\
                                     & \textbf{Count} & \textbf{Comp} & \textbf{Avg} & \textbf{Count} & \textbf{Local} & \textbf{Avg} & \textbf{Exist} & \textbf{Count} & \textbf{Local} & \textbf{Comp} & \textbf{Temp} & \textbf{Avg} & \\ \hline
        FCNLSTM~\cite{fayek2020temporal}                     & 70.45 & 66.22 & 68.88 & 63.89 & 46.74 & 55.21 & 82.01 & 59.34 & 46.28 & 62.15 & 47.33 & 60.06 & 60.34 \\
    BiLSTM~\cite{zhou2016attention}                   & 70.35 & 47.92 & 62.05 & 64.64 & 64.33 & 64.48 & 78.39 & 56.91 & 45.85 & 53.09 & 49.76 & 57.10 & 59.92 \\
    HCAttn~\cite{lu2016hierarchical}                   & 70.25 & 54.91 & 64.57 & 64.05 & 66.37 & 65.22 & 79.10 & 59.97 & 49.51 & 55.25 & 56.43 & 60.19 & 62.30 \\
    MCAN~\cite{yu2019deep}                       & 77.50 & 55.24 & 69.25 & 71.56 & 70.93 & 71.24 & 80.40 & 64.91 & 54.48 & 57.22 & 47.57 & 61.58 & 65.49 \\
    PSAC~\cite{li2019beyond}                       & 75.64 & 66.06 & 72.09 & 68.64 & 69.79 & 69.22 & 77.59 & 63.42 & 55.02 & 61.17 & 59.47 & 63.52 & 66.54 \\
    HME~\cite{fan2019heterogeneous}                         & 74.76 & 63.56 & 70.61 & 67.97 & 69.46 & 68.76 & 80.30 & 63.19 & 53.18 & 62.69 & 59.83 & 64.05 & 66.45 \\
    HCRN~\cite{le2020hierarchical}                       & 68.59 & 50.92 & 62.05 & 64.39 & 61.81 & 63.08 & 54.47 & 53.38 & 41.53 & 52.11 & 47.69 & 50.26 & 55.73 \\
                                     AVSD~\cite{schwartz2019simple}                       & 72.41 & 61.90 & 68.52 & 67.39 & 74.19 & 70.83 & 81.61 & 63.89 & 58.79 & 61.52 & 61.41 & 65.49 & 67.44 \\
    ST-AVQA~\cite{li2022learning}              & 78.18 & 67.05 & 74.06 & 71.56 & 76.38 & 74.00 & 81.81 & 70.80 & 64.51 & {66.01} & 63.23 & 69.54 & 71.52 \\
    COCA~\cite{lao2023coca}                       & 79.35 & 67.68 & 75.42 & 75.10 & 75.43 & 75.23 & \ul{83.50} & 66.63 & 69.72 & 64.12 & 65.57 & 69.96 & 72.33 \\
    PSTP-Net~\cite{li2023progressive}                   & 73.97 & 65.59 & 70.91 & 77.15 & 77.36 & 77.26 & 76.18 & 72.23 & 71.80 & \textbf{71.79} & 69.00 & 72.57 & 73.52 \\
    LAVISH~\cite{lin2023vision}                   & 82.09 & 65.56 & 75.97 & 78.98 & 81.43 & 80.22 & 81.71 & 75.51 & 66.13 & 63.77 & 67.96 & 71.26 & 74.46 \\
    APL~\cite{li2024object}                         & 82.40 & \textbf{70.71} & 78.09 & 76.52 & 82.74 & 79.69 & 82.99 & 73.29 & 66.68 & 64.76 & 65.95 & 70.96 & 74.53 \\
    TSPM$^{\S}$~\cite{li2024boosting}                       & 84.07 & 64.65 & 76.91 & 82.29 & 84.90 & 83.61 & 82.19 & {76.21} & \ul{71.85} & 65.76 & \textbf{71.17} & \ul{73.51} & 76.79 \\
    VideoLLaMA-2$^{\ddagger}$~\cite{cheng2024videollama} & 79.44 & 52.46 & 69.64 & 81.30 & 82.93 & 82.11 & 77.00 & 77.69 & 63.44 & 59.46 & 64.71 & 68.98 & 72.56 \\
    AVMoE~\cite{cheng2024avmoe}                    & -- & -- & 77.60 & -- & -- & 82.70 & -- & -- & -- & -- & -- & 71.90 & 75.70 \\
    MAVEN~\cite{ma2025fortisavqa}                  & 79.44 & 54.10 & 72.79 & 80.49 & \textbf{93.50} & \textbf{86.99} & \textbf{87.00} & 73.85 & 66.67 & 54.95 & 68.24 & 69.94 & 74.60 \\
    QA-TIGER$\dagger$$^{\S}$\cite{2025tiger}  & \ul{85.25}  & \ul{68.01}  & \ul{78.90}  & \ul{84.71}  & 86.29  & 85.51 & 82.49  & \ul{78.74}  & 71.41  & 63.94  & 68.86  & 73.35  &  \ul{77.56} \\ \hline
     \textbf{SHRIKE}$^{\S}$ \graycell & \textbf{85.84} \graycell & \ul{68.01} \graycell & \textbf{79.27} \graycell & \textbf{85.30} \graycell & \ul{86.86} \graycell & \ul{86.09} \graycell& {80.26} \graycell & \textbf{79.05} \graycell & \textbf{72.72} \graycell & \ul{66.76} \graycell & \ul{69.83} \graycell & \textbf{74.00} \graycell & \bf \graycell {78.14} \\
    \bottomrule
    \end{tabular}
}
\end{adjustbox}
\end{table*}

\begin{table}[t]
\renewcommand{\arraystretch}{1.4}
\renewcommand{\tabcolsep}{1.2mm}
\footnotesize
\centering
\caption{Comparison with state-of-the-art methods on the Music-AVQA v2 dataset. The upper table shows results on the biased test set, and the bottom table shows results on the balanced test set.}
\label{tab:music-avqa-v2}
\begin{subtable}{\linewidth}
\centering
\begin{adjustbox}{width=\linewidth}
\begin{tabular}{C{10mm}|c|l|cccc}
\toprule
\textbf{Test} & \textbf{Training} & \textbf{Method} & \textbf{A-QA} & \textbf{V-QA} & \textbf{AV-QA} & \textbf{Avg} \\
\hline
\multirow{10}{*}{\small Bias}
& \multirow{4}{*}{Bias}        & ST-AVQA~\cite{li2022learning}      & {76.86}     & 77.70          & 69.59          & 73.07          \\
&                              & LAVISH~\cite{lin2023vision}         & 76.73          & {80.96}     & {70.80}     & {74.59}     \\
&                              & QA-TIGER\dag\cite{2025tiger}         & \textbf{78.94} & \ul{84.99} & \ul{72.63} & \ul{77.08} \\
&                              & \textbf{SHRIKE}                      & \ul{78.39} & \textbf{86.09} & \textbf{72.73} & \textbf{77.33} \\
\cline{2-7}
& \multirow{6}{*}{Balance}     & ST-AVQA~\cite{li2022learning}      & 76.18          & 77.20          & 67.96          & 71.92          \\
&                              & LAVISH~\cite{lin2023vision}         & 75.56          & 80.83          & 69.27          & 73.51          \\
&                              & LAST~\cite{liu2024tackling}         & {77.10}     & 82.99          & 70.86          & 75.24          \\
&                              & LAST-Att~\cite{liu2024tackling}     & 77.29 & {83.47}     & {71.05}     & {75.45}     \\
&                              & QA-TIGER\dag\cite{2025tiger}        & \ul{78.39} & \ul{85.93} & \textbf{71.68} & \ul{76.71} \\
&                              & \textbf{SHRIKE}                     & \textbf{78.82} & \textbf{87.40} & \ul{71.30} & \textbf{76.97} \\
\bottomrule
\end{tabular}
\end{adjustbox}
\end{subtable}

\vspace{3.5mm}

\begin{subtable}{\linewidth}
\centering
\begin{adjustbox}{width=\linewidth}
\begin{tabular}{C{10mm}|c|l|cccc}
\toprule
\textbf{Test} & \textbf{Training} & \textbf{Method} & \textbf{A-QA} & \textbf{V-QA} & \textbf{AV-QA} & \textbf{Avg} \\
\hline
\multirow{10}{*}{\small Balance}
& \multirow{4}{*}{Bias}        & ST-AVQA~\cite{li2022learning}   & {73.34}     & 76.82          & 64.51          & 69.40          \\
&                              & LAVISH~\cite{lin2023vision}        & 73.14          & {79.70}     & {65.01}     & {70.39}     \\
&                              & QA-TIGER\dag\cite{2025tiger}       & \textbf{78.12} & \ul{84.84} & \ul{68.06} & \ul{74.35} \\
&                              & \textbf{SHRIKE}                    & \ul{77.33} & \textbf{85.90} & \textbf{68.19} & \textbf{74.44} \\
\cline{2-7}
& \multirow{6}{*}{Balance}     & ST-AVQA~\cite{li2022learning}   & 75.50          & 77.67          & 66.32          & 71.02          \\
&                              & LAVISH~\cite{lin2023vision}        & 76.15          & 81.32          & 68.28          & 73.18          \\
&                              & LAST~\cite{liu2024tackling}        & 78.08          & 83.29          & 69.72          & 74.85          \\
&                              & LAST-Att~\cite{liu2024tackling}    & {78.56}     & {84.07}     & {70.30} & {75.44}     \\
&                              & QA-TIGER\dag\cite{2025tiger}       & \textbf{79.31} & \ul{86.28} & \ul{70.11} & \ul{76.08} \\
&                              & \textbf{SHRIKE}                    & \ul{78.91} & \textbf{87.44} & \textbf{70.36} & \textbf{76.45} \\
\bottomrule
\end{tabular}
\end{adjustbox}
\end{subtable}
\end{table}

\begin{figure*}[!htbp]
    \centering
    \includegraphics[width=1
    \linewidth]{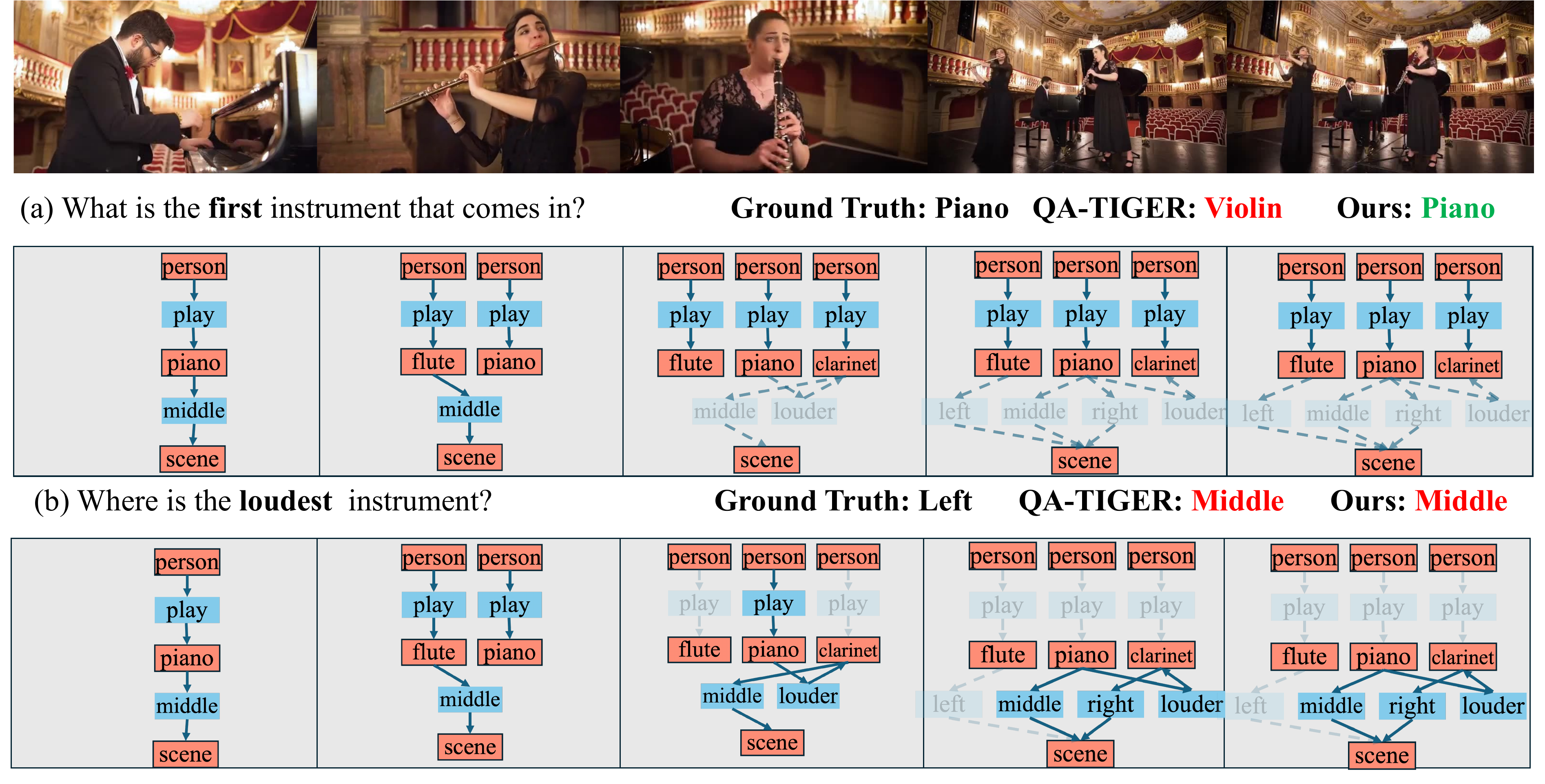}
    \vspace{-3mm}
    \caption{Visualized relationship triplet selection results. In the given example, we convert all the relationship triplets into a scene graph and highlight the triplets selected by our method.}
    \label{Fig:REL}
\end{figure*}

\noindent\textbf{Metrics.} Following the standard evaluation protocol of the original benchmark~\cite{li2022learning}, we adopt top-1 answer accuracy as our evaluation metric, where a prediction is regarded as correct only when it exactly matches the ground-truth answer. As the candidate answers form a fixed vocabulary, this is equivalent to selecting the answer with the highest predicted probability and comparing it against the annotation. To provide a comprehensive picture of the model's behavior across modalities, we report accuracy separately on the three major question groups, namely \emph{Audio QA}, \emph{Visual QA}, and \emph{Audio-Visual QA}, and we also summarize the overall average accuracy computed over all questions. In addition, for a finer-grained analysis, we further break the results down into per-category accuracy (\emph{e.g.}, counting, location, and comparative questions), consistent with the evaluation setting used in prior work~\cite{lao2023coca,li2023progressive,li2024object,li2024boosting,2025tiger}. Reporting these complementary metrics allows us to examine not only the overall performance but also the specific question types on which the improvements of our method are most pronounced.




\subsection{Implementation Details}
\label{sec:details}
Our model is implemented in PyTorch and trained on a single NVIDIA RTX 3090 GPU for a total of 25 epochs with a batch size of 32. We adopt a two-stage training schedule: during the first 10 epochs, we train only the Multi-Modal Scene Graph Decoder; afterwards, we freeze its parameters and train the remaining modules during the subsequent 15 epochs, so that the predicted scene graphs remain independent of the specific question. The input videos are uniformly sampled at a rate of 1 fps. For feature extraction, audio features are obtained with the pre-trained VGGish model~\cite{vgg2017}, whereas the visual frames and the question text are encoded with CLIP-ViTL/14~\cite{radford2021learning}. To keep the representations consistent across modalities, all extracted features are linearly projected to a common dimensionality of
512. For optimization, we use the Adam optimizer with an initial learning rate of $1e^{-4}$, which is decayed by a factor of 0.1 every 9 epochs. During scene graph extraction, each video is evenly divided into 10 segments and a separate scene graph is generated for every segment. Unless otherwise specified, we set $top_{k} =10$ in the proposed modules. More details are provided in the supplementary.

\subsection{Quantitative Results}

We compare SHRIKE against state-of-the-art approaches on both MUSIC-AVQA and Music-AVQA v2.0, and the detailed results are summarized in Table~\ref{tab: music-avqa} and Table~\ref{tab:music-avqa-v2}, respectively.

{\bf For MUSIC-AVQA~\cite{li2022learning}}, SHRIKE attains the best overall average accuracy of 78.14\%, establishing a new state of the art. It surpasses the strongest competitor QA-TIGER~\cite{2025tiger} (77.56\%) by 0.58\%, and outperforms earlier methods such as TSPM~\cite{li2024boosting} (76.79\%) and LAVISH~\cite{lin2023vision} (74.46\%) by even larger margins. Moreover, SHRIKE achieves the best average accuracy across all three modality groups, namely Audio QA (79.27\%), Visual QA (86.09\%), and Audio-Visual QA (74.00\%), consistently outperforming QA-TIGER (78.90\%, 85.51\%, and 73.35\%, respectively). The advantage is especially pronounced on questions that require joint audio-visual reasoning: SHRIKE delivers clear gains over QA-TIGER on the Localization (72.72\% vs 71.41\%, $+1.31\%$), Temporal (69.83\% vs 68.86\%, $+0.97\%$), and Comparative (66.76\% vs 63.94\%, $+2.82\%$) subsets. These gains demonstrate that the explicit relational structure provided by our multi-modal scene graph is particularly effective for localizing and comparing question-relevant cross-modal cues. In addition, SHRIKE also compares favorably with more recent approaches. In the evaluation reported by MAVEN~\cite{ma2025fortisavqa}, VideoLLaMA-2 attains an overall accuracy of 72.56\%, while MAVEN achieves 74.60\%. AVMoE~\cite{cheng2024avmoe} reports modality-level accuracies of 77.60\%, 82.70\%, and 71.90\% for Audio QA, Visual QA, and Audio-Visual QA, respectively, resulting in an overall accuracy of 75.70\%. All these results remain below the 78.14\% achieved by SHRIKE, further demonstrating the effectiveness of task-specific structured supervision for fine-grained audio-visual reasoning.

{\bf For MUSIC-AVQA v2.0}, SHRIKE again achieves the best overall average accuracy under all four training/test configurations, consistently ranking first (77.33\% vs 77.08\%, 76.97\% vs 76.71\%, 74.44\% vs 74.35\%, 76.45\% vs 76.08\%) and thus maintaining its lead on both the biased and the balanced splits. The improvement is especially notable on Visual QA, where SHRIKE consistently surpasses QA-TIGER by roughly one to one-and-a-half points across all settings (\emph{e.g.}, 87.40\% vs 85.93\% and 87.44\% vs 86.28\%), reflecting the benefit of incorporating structured visual cues into the reasoning process. Taken together, the consistent improvements on both benchmarks highlight the strong and robust reasoning capabilities of our proposed method.

\subsection{Qualitative Results}
Figure~\ref{Fig:REL} presents a side-by-side comparison between the predictions of QA-TIGER~\cite{2025tiger} and those of our SHRIKE on the same audio-visual input, along with the multi-modal scene graphs generated for that example. In Figure~\ref{Fig:REL}(a), for the question “What is the first instrument that comes in?”, our model correctly recognizes the piano and preserves the relevant relationship triplets, such as $\langle person, play, piano\rangle$ and $\langle piano, louder than, flute\rangle$, which provide explicit structural context and thereby improve the model's structural visual understanding. In Figure~\ref{Fig:REL}(b), in contrast, the model produces an incorrect answer. Although it still retains triplets describing position and loudness, the absence of explicit comparative relations with the other instruments during scene graph generation ultimately misleads the reasoning process. We attribute this failure mainly to the long-tail distribution of audio loudness, which makes it difficult for the model to reliably estimate the relative volume of less salient instruments and consequently causes some key audio triplets to be missed.

\subsection{Ablation Studies}

\noindent\textbf{Multimodal Scene Graph and KAN-based MoE.} To validate the contribution of each designed component, we conduct ablation experiments on our two core components, \emph{i.e.}, the Temporal Integration with Gaussian KAN-based Experts and the Scene Graph Decoder, and report the results in Table~\ref{tab:ablation}. Starting from the baseline, incorporating the Multi-Modal Scene Graph alone improves the average accuracy by 1.13\% (77.56\% vs. 76.43\%), which confirms that the explicit structural cues provided by the scene graph are beneficial for audio-visual reasoning. On top of this, when the KAN-based experts are further introduced into the Temporal Integration module, the average accuracy increases by an additional 0.58\% (78.14\% vs. 77.56\%), demonstrating that the two components are complementary and that each brings a consistent and independent gain.

\begin{figure}[t]
\centering
\begin{minipage}{\linewidth}
    \centering
    \captionof{table}{Ablation study of the proposed framework. M$^2$SG denotes the Multi-Modal Scene Graph Module.}
    \label{tab:ablation}
    \footnotesize
    \renewcommand{\arraystretch}{1.2}
    \begin{tabular*}{\linewidth}{@{\extracolsep{\fill}}c|c|c c c c@{}}
    \toprule
    \multicolumn{1}{c|}{\textbf{M$^2$SG}} & \textbf{KAN} & \textbf{A-QA} & \textbf{V-QA} & \textbf{AV-QA} & \textbf{Avg} \\
    \midrule
      &         & 77.96 & 83.44 & 72.50 & 76.43 \\
    \checkmark &       & 78.52 & 86.13 & 73.18 & 77.56 \\
      & \checkmark & 78.09 & 85.57 & 73.21 & 77.33 \\
    \checkmark & \checkmark & \textbf{79.27} & \textbf{86.09} & \textbf{74.00} & \textbf{78.14} \\
    \bottomrule
    \end{tabular*}
\end{minipage}
\end{figure}

\captionsetup[sub]{skip=2pt}
\begin{figure*}[!tbp]
    \centering
    \begin{minipage}[b]{1\textwidth}
        \centering
        \includegraphics[width=1\textwidth,trim=0 0 0 10,clip]{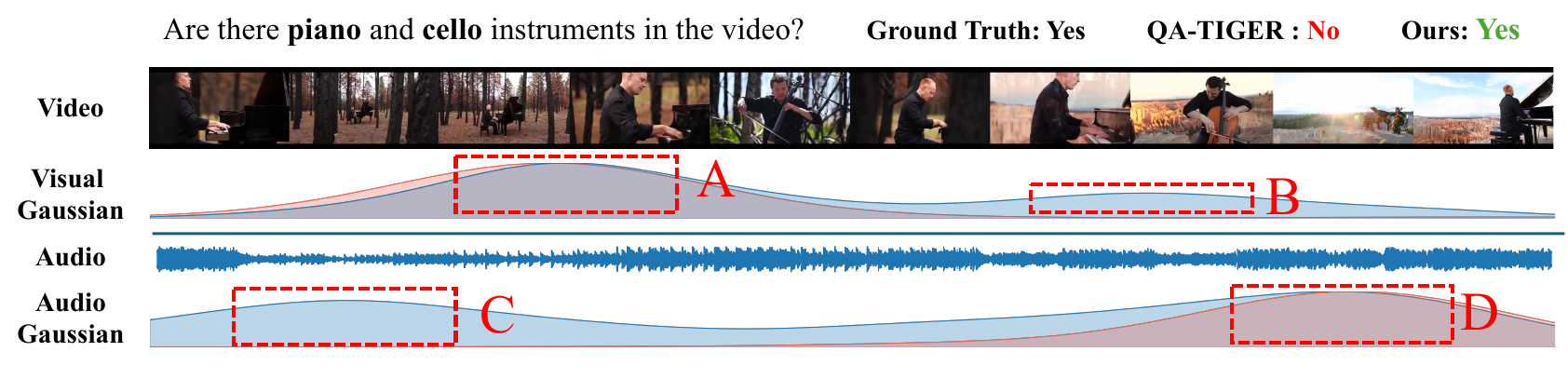}
        \vspace{-5mm}
        \subcaption{Visual Question}
        \label{qualitative1}
    \end{minipage}\vspace{2mm}
    \begin{minipage}[b]{1\textwidth}
        \centering
        \includegraphics[width=\linewidth,trim=0 0 0 8,clip]{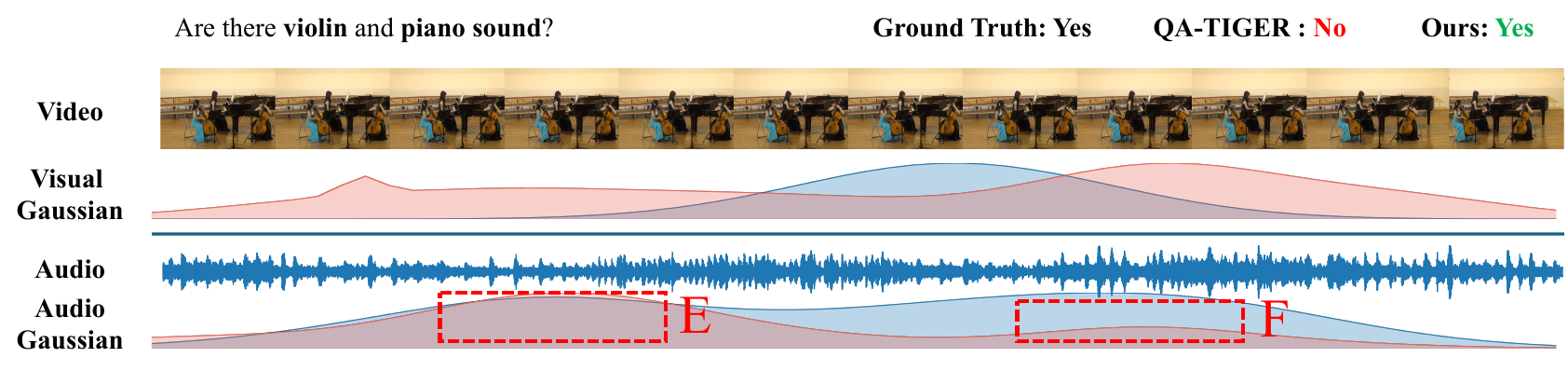}
        \vspace{-5mm}
        \subcaption{Audio Question}
        \label{qualitative2}
    \end{minipage}

    \begin{minipage}[b]{1\textwidth}
        \centering
        \includegraphics[width=1\textwidth,trim=0 0 0 8,clip]{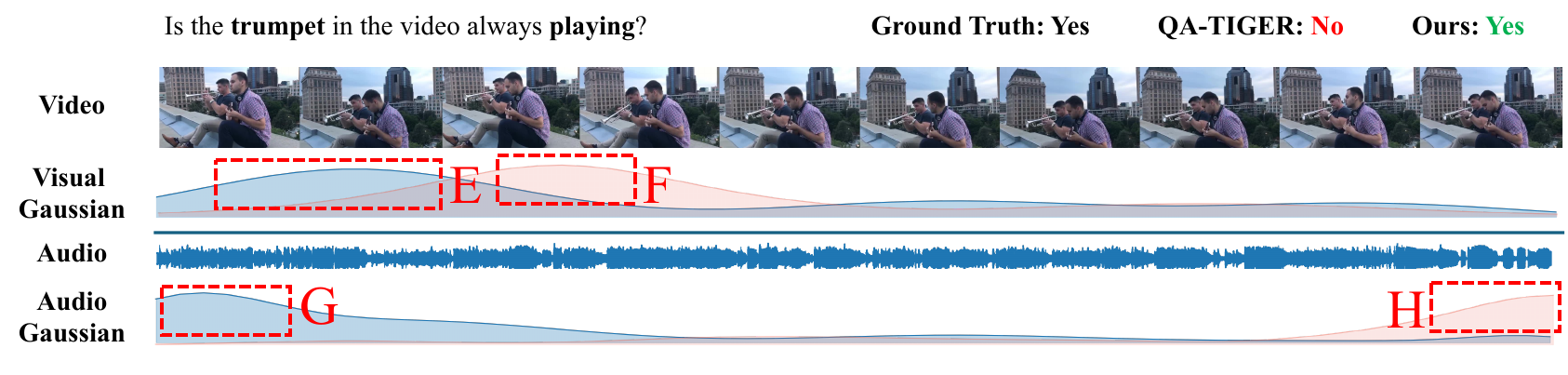}
        \vspace{-5mm}
        \subcaption{Audio-Visual Question}
        \label{qualitative3}
    \end{minipage}
    \vspace{-6mm}
    \caption{Comparison of key temporal segment capture between SHRIKE and QA-TIGER. \textcolor{blue}{Blue} wavy lines represent SHRIKE and
    \textcolor{red}{Red} wavy lines represent QA-TIGER.\qms{the font is too large!}}
    \label{Fig:visuailzation}
    \vspace{-3mm}
\end{figure*}

\begin{figure}[t]
\begin{minipage}{\linewidth}
    \centering
    \includegraphics[width=\linewidth,trim=0 15 0 12,clip]{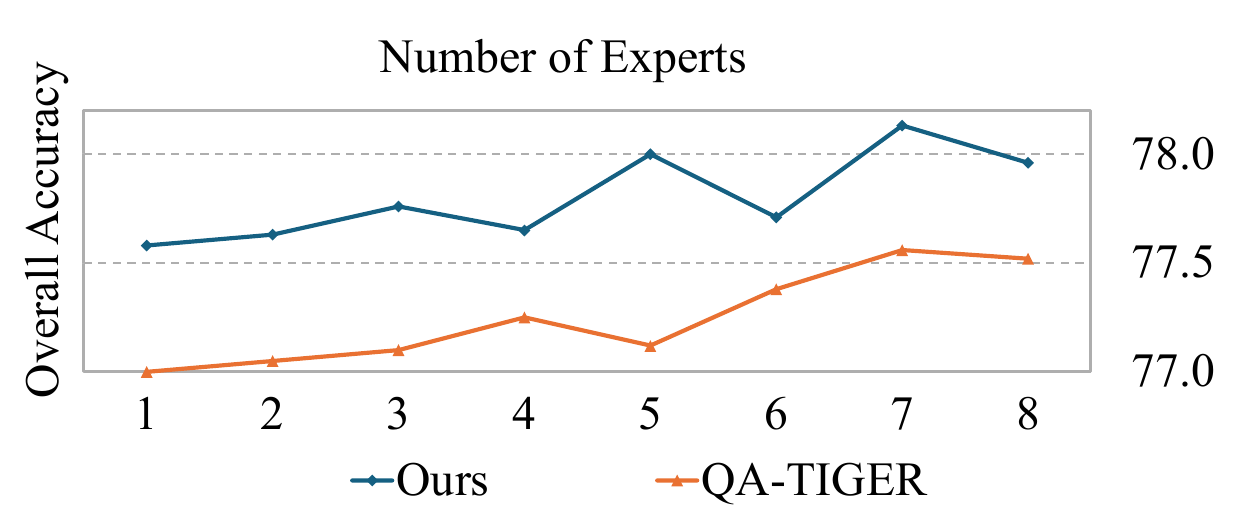}
    \vspace{-2mm}
    \captionof{figure}{Model performance with numbers of KAN experts.}
    \label{Fig:CKAN}
\end{minipage}

\end{figure}

\noindent\textbf{Selection of KAN Experts.}
\label{sec:KAN}
We further investigate how the number of KAN experts affects the final performance, and the corresponding results are shown in Figure \ref{Fig:CKAN}. Overall, the accuracy of the model improves gradually as the number of experts increases, which suggests that employing more experts helps to capture a richer set of visual and audio cues. Moreover, the rate of improvement is not uniform but tends to accelerate when more experts are used ({\it e.g.}, the fastest growth occurs when the number of experts increases from 6 to 7). In addition, across the entire range of settings, our method consistently outperforms QA-TIGER~\cite{2025tiger}, which further validates the superiority of the proposed KAN-based temporal integration over its MLP-based counterpart.

\begin{figure}[t]
\begin{minipage}{\linewidth}
    \centering
    \includegraphics[width=\linewidth,trim=0 0 0 12,clip]{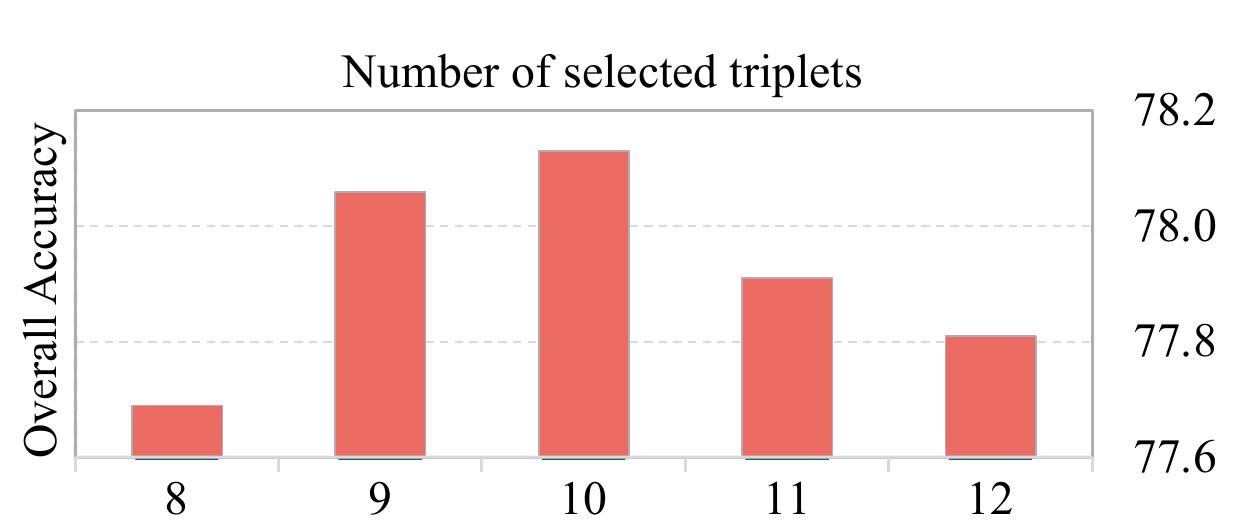}
    \vspace{-2mm}
    \captionof{figure}{Model performance with different $top_{k}$ values.}
    \label{Fig:Topk}
\end{minipage}
\end{figure}

\noindent\textbf{Selection of Relationship Triplets.}
\label{sec:REL}
We examine how the number of relationship triplet sets selected per frame, denoted by $top_{k}$, influences the overall performance, and the corresponding results are reported in Figure \ref{Fig:Topk}. Even with a relatively small budget, the selection strategy already proves effective: when $top_{k}$ is set to 8, the model reaches an accuracy of 77.69\%, which already surpasses the state-of-the-art QA-Tiger model. As $top_{k}$ is further increased to $10$, the accuracy continues to improve and reaches 78.14\%. This trend clearly highlights the advantage of explicitly selecting the most relevant triplets for the subsequent reasoning.

\begin{table}[t]
\centering
\footnotesize
\caption{Ablation study on the various number of video segments.}
\label{tab:segment}
\renewcommand{\arraystretch}{1.2}
\setlength{\tabcolsep}{3pt}
\begin{tabular*}{\columnwidth}{@{\extracolsep{\fill}}l|ccc|c}
\toprule
\textbf{Number of Segments} & \textbf{A-QA} & \textbf{V-QA} & \textbf{AV-QA} & \textbf{Average} \\
\midrule
5 segments  & 68.00 & 60.44 & 58.73 & 60.61 \\
12 segments & 66.00 & 64.84 & 59.26 & 61.82 \\
10 segments & 66.00 & 61.54 & 61.90 & \textbf{62.42} \\
\bottomrule
\end{tabular*}
\end{table}

\noindent\textbf{Number of video segments.}
We conduct an ablation study to analyze the effect of the number of video segments, using a subset of 500 sampled videos for efficiency. Specifically, we construct scene graphs by dividing each video into 5, 10, and 12 segments, respectively, and then evaluate the resulting annotations on the downstream task under identical settings. As shown in Table~\ref{tab:segment}, dividing each video into 10 segments yields the best overall performance, which suggests that this granularity offers a suitable trade-off between temporally disentangling distinct events and preserving sufficient contextual information within each segment.

\noindent\textbf{Robustness to Encoder Backbones.}
To verify that the advantage of our method does not depend on a particular feature extractor, we further evaluate all models under different encoder backbones. As shown in Table~\ref{dif encoder}, SHRIKE consistently achieves the best performance across all encoder configurations, which confirms the robustness and generality of the proposed framework.
\begin{table}[t]
\centering
\caption{Performance with different encoders serving both visual and text feature extraction.}
\label{dif encoder}
\setlength{\tabcolsep}{6pt}
\renewcommand{\arraystretch}{1.12}
\resizebox{\columnwidth}{!}{%
\begin{tabular}{l c c c c c}
\toprule
Method & CLIP Encoder & A-QA & V-QA & AV-QA & Avg \\
\midrule
PSTP-Net\cite{li2023progressive} & B/32 & 70.91 & 77.26 & 72.57 & 73.52 \\
TSPM\cite{li2024boosting}     & B/32 & 76.91 & 81.92 & 72.57 & 75.81 \\
QA-TIGER\cite{2025tiger}     & B/32 & 76.23 & \textbf{84.10} & 72.14 & 76.03 \\
\textbf{Ours} & B/32 & \textbf{78.15} & 83.86 & \textbf{72.23} & \textbf{76.36} \\
\bottomrule
\end{tabular}
}
\end{table}


\subsection{Analysis of Temporal Integration} 
To better understand how the proposed temporal integration behaves, we provide a qualitative analysis based on the learned Gaussian distributions over time, as illustrated in Figure~\ref{qualitative1}, Figure~\ref{qualitative2}, and Figure~\ref{qualitative3}. In each case, the blue curves correspond to the temporal weights learned by our KAN-based experts, while the red curves correspond to those of the MLP-based experts in QA-TIGER~\cite{2025tiger}, which allows a direct comparison of how the two designs attend to question-relevant temporal segments.

\noindent\textbf{Visual Question.}  
As shown in Figure \ref{qualitative1}, both QA-TIGER~\cite{2025tiger} and SHRIKE are able to identify the appearance of the piano in region A from visual cues, as reflected by high values of both blue and red distributions in that region. However, QA-TIGER~\cite{2025tiger} fails to detect the cello in region B of the visual stream. QA-TIGER only localizes one instrument in the visual stream and another in the audio stream, and is thus unable to align two modalities. In contrast, by incorporating structured visual information, our SHRIKE successfully detects both instruments across both modalities at the same time (regions A and B for the visual stream, and regions C and D for the audio stream), and consequently arrives at the correct answer.

\noindent\textbf{Audio Question.} 
As shown in Figure~\ref{qualitative2}, for a question that mainly relies on the audio modality, both our SHRIKE and QA-TIGER are able to detect the presence of the first instrument in region E of the audio cues. The key difference appears in region F: leveraging the powerful modeling capability of KAN, our method successfully identifies the appearance of the second instrument in this region, whereas QA-TIGER overlooks it and therefore produces an incorrect answer. This comparison illustrates that the KAN-based experts are more sensitive to subtle audio events that are easily missed by the MLP-based counterpart.

\noindent\textbf{Audio-Visual Question}.
As shown in Figure~\ref{qualitative3}, correctly answering this question requires the model to reason jointly over both audio and visual cues. Although both QA-TIGER~\cite{2025tiger} and our method are able to detect the trumpet in the visual stream, our method localizes this cue earlier than QA-TIGER (region E vs. region F), which is important for judging whether the instrument is being played throughout the clip. Similarly, on the audio side, our method also captures the relevant cue earlier (region G vs. region H). Thanks to this tighter alignment between the audio and visual evidence, our model is able to produce the correct answer, which demonstrates its stronger audio-visual reasoning ability. Overall, across all of above cases, blue curves exhibit clearer and earlier peaks than red ones, which indicates that the KAN-based temporal integration learns temporal weights that are both more discriminative and better aligned with the truly relevant segments than those learned by the MLP-based counterpart. This observation highlights the stronger expressive power of the KAN experts for modeling complex cross-modal temporal dependencies.

\begin{table*}[t]
\centering
\small
\caption{Comparison using scene graph annotations generated by different MLLMs and direct QA performance of Qwen3 Omni on the MUSIC-AVQA dataset.}
\label{tab:mlm_comparison}
\renewcommand{\arraystretch}{1.2}
\begin{tabular*}{\textwidth}{@{\extracolsep{\fill}} l|l ccc|c}
\toprule
\textbf{Dataset} & \textbf{Method} & \textbf{A-QA} & \textbf{V-QA} & \textbf{AV-QA} & \textbf{Average} \\
\midrule
MUSIC-AVQA (subset) & Ours w/ Qwen3 Omni  & 72.56 & 78.49 & 69.33 & \textbf{72.33} \\
MUSIC-AVQA (subset) & Ours w/ Qwen2.5 Omni & 69.09 & 78.29 & 66.03 & 69.76 \\
\midrule
MUSIC-AVQA (full) & Only Qwen3 Omni & 59.81 & 81.18 & 62.30 & 67.29 \\
MUSIC-AVQA (full) & Ours w/ Qwen3 Omni & 79.27 & 86.09 & 74.00 & \textbf{78.14} \\
\bottomrule
\end{tabular*}
\end{table*}

\begin{table}[t]
\centering
\footnotesize
\caption{Generalization and transferability on related audio-visual understanding tasks.}
\label{tab:general}
\renewcommand{\arraystretch}{1.2}
\begin{tabular*}{\columnwidth}{@{\extracolsep{\fill}} l l c c c c @{}}
\toprule
\textbf{Task} & \textbf{Data} & \textbf{Metric}
& \textbf{Base.} & \textbf{Ours} & \textbf{$\Delta$} \\
\midrule
Event loc. & AVE & Acc.
& 78.48 & 78.61 & \textcolor{red}{+0.13} \\
\midrule
\multirow{2}{*}{Seg.}
& \multirow{2}{*}{AVSBench}
& mIoU
& 51.16 & 51.74 & \textcolor{red}{+0.58} \\
& & $F_{\mathrm{score}}$
& 63.39 & 63.62 & \textcolor{red}{+0.23} \\
\midrule
QA & WorldSense & Acc.
& 29.08 & 31.21 & \textcolor{red}{+2.13} \\
\bottomrule
\end{tabular*}
\end{table}

\subsection{Discussion}
\noindent\textbf{Choice of MLLM for Multimodal Scene Graph Annotation.}
To determine which MLLM is most suitable for automatic scene graph annotation, we compare Qwen3 Omni-30A3B~\cite{xu2025qwen3} and Qwen2.5 Omni-7B~\cite{xu2025qwen2omni}. Specifically, we use each model to generate scene graph annotations for the same 2{,}000 annotated videos under identical settings, and then evaluate the quality of the generated annotations through downstream performance. As shown in Table~\ref{tab:mlm_comparison}, Qwen3 Omni-30A3B achieves better performance, which indicates its stronger capability in producing more accurate and reliable scene graph annotations. Based on this observation, we adopt Qwen3-Omni-30A3B as the annotation model for the construction of our scene graph supervision.\\
\noindent\textbf{Performance of Qwen3 Omni on the MUSIC-AVQA Dataset.} To estimate how much task-specific supervision Qwen3 Omni implicitly injects through the pseudo scene graph annotations, we additionally evaluate it directly on the MUSIC-AVQA QA task, using the Qwen3 Omni-30A3B model for testing. As shown in Table~\ref{tab:mlm_comparison}, when applied directly to question answering, Qwen3 Omni achieves only 67.29\%, which is far below that of dedicated supervised audio–visual QA models. This result suggests that the scene graph annotations mainly provide coarse semantic cues for training, rather than acting as near-oracle labels that would trivially solve the task.\\
\noindent\textbf{Generalization and Transferability.}
We further assess the generalization ability of our framework by applying it to several related audio-visual understanding tasks, with the results summarized in Table~\ref{tab:general}. On the AVE dataset~\cite{tian2018audio}, M$^2$SG achieves 78.61\% accuracy on audio-visual event localization, which slightly surpasses CACE-Net~\cite{he2024cace} (78.48\%). On AVSBench, our framework also improves multi-source audio-visual segmentation, raising the mIoU from 51.16 to 51.74 and the $F_{\text{score}}$ from 63.39 to 63.62. Beyond these tasks, an important property of our framework is that it is not tied to any fixed object or relation vocabulary. Although the categories adopted here are specifically designed for Music-AVQA, the same annotation pipeline can be readily adapted to other domains simply by redefining the underlying vocabulary. To validate this scalability, we further test on WorldSense~\cite{hong2025worldsense}: after filtering out overly long videos, 940 videos (1,748 samples) are retained, and by normalizing the candidates into 36 objects and 13 relations, our method achieves 31.21\% accuracy, outperforming QA-TIGER by 2.13\% (29.08\%). These results clearly demonstrate the scalability and transferability of the proposed scene graph framework across diverse audio-visual scenarios.
\begin{table}[t]
    \centering
    \footnotesize
    \caption{Comparison of MLP and KAN without the Relationship Triplet Selection module.}
    \label{tab:kan_mlp_no_rts}
    \renewcommand{\arraystretch}{1.2}
    \setlength{\tabcolsep}{3pt}
    \begin{tabular*}{\columnwidth}{@{\extracolsep{\fill}} l|cccc}
        \toprule
        \textbf{Method} & \textbf{A-QA} & \textbf{V-QA} & \textbf{AV-QA} & \textbf{Average} \\
        \midrule
        SHRIKE w/ MLP & 79.58 & 85.88 & 73.10 & 77.63 \\
        SHRIKE w/ KAN & 77.96 & 86.09 & 73.88 & \textbf{77.84} \\
        \bottomrule
    \end{tabular*}
\end{table}

\noindent
\textbf{Disentangling relation selection from temporal modeling.}
We now clarify the distinction between the Relationship Triplet Selection module and the Gaussian KAN-based temporal modeling, as the two operate at different levels. The Relationship Triplet Selection module only filters the query-relevant triplets within each individual segment and does not compare different segments over time; consequently, it improves the relevance of the per-segment relation features but does not, by itself, perform any temporal localization. In contrast, the Gaussian KAN-based Experts operate on the entire sequence of segment-level features, assigning time-wise importance and aggregating evidence across segments. The two mechanisms are therefore complementary rather than redundant: the former refines local relational cues within each segment, while the latter captures the temporal dependencies that span across segments. Thus, we further conduct an additional ablation study in which we remove the Relationship Triplet Selection module and directly compare the MLP-based and KAN-based variants, with the results reported in Table~\ref{tab:kan_mlp_no_rts}, demonstrating the advantage of the KAN.
\\

\section{Conclusion}
\label{Conclusion}
In this paper, we presented the first framework that integrates multi-modal scene graphs with KAN for audio–visual question answering. Specifically, we designed a Scene Graph Decoder to extract relation triplets from fused audio–visual features, providing explicit structural cues for reasoning, and introduced a KAN-based Mixture-of-Experts module to enhance the expressive power of temporal integration. Extensive experiments on multiple benchmark datasets demonstrate its effectiveness, generalizability, and superiority over existing methods. In the future, we will extend our method in the human-robot interaction.

\bibliographystyle{IEEEtran}
\bibliography{main}

\vspace{-5mm}

\begin{IEEEbiography}[{\includegraphics[width=1in,height=1.25in]{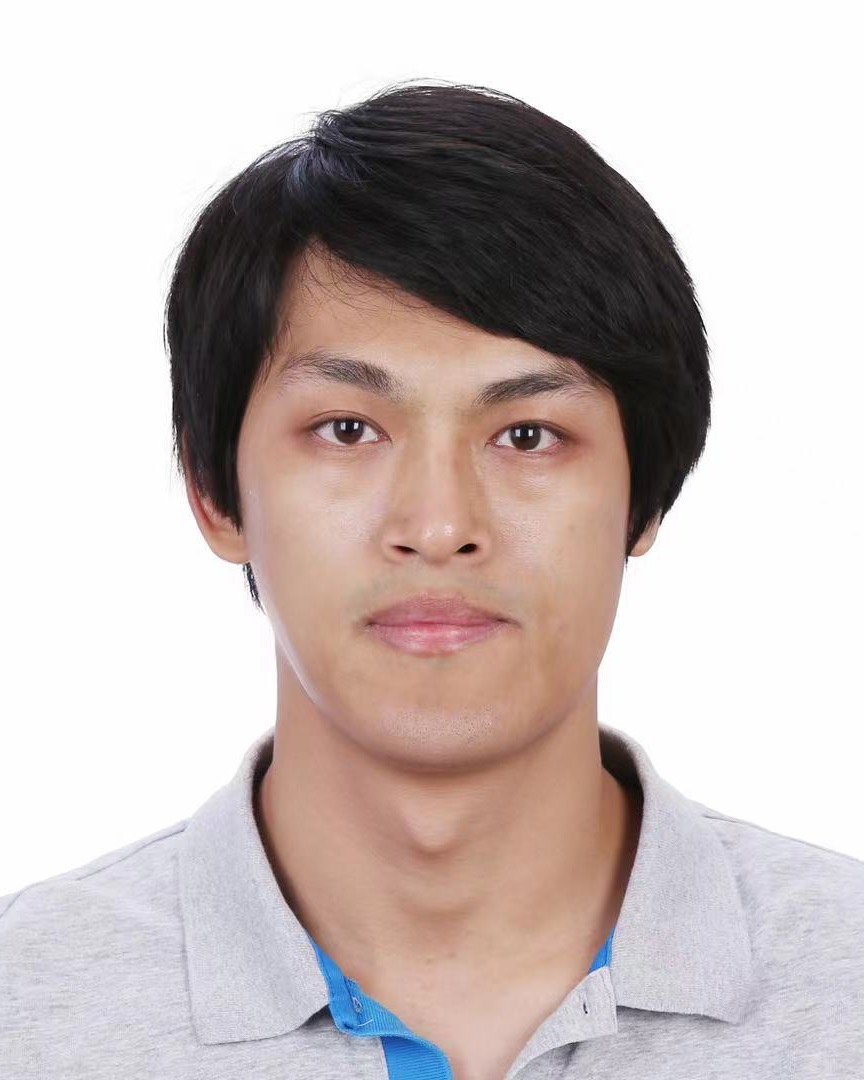}}] {Mengshi Qi} (Member, IEEE) is currently a Professor with the Beijing University of Posts and Telecommunications, Beijing, China. He received Ph.D. degree in computer science from Beihang University, Beijing, China, in 2019. His research interests include machine learning and computer vision, especially scene understanding, 3D reconstruction, and multimedia analysis. He has published more than 50 papers in top journals (such as IEEE TPAMI, TIP, TMM, TCSVT, TIFS) and top conferences (such as IEEE CVPR, ICCV, ECCV, ACM Multimedia, AAAI, NeurIPS). He also has served as Guest Editor of IEEE TMM, Area Chair of IEEE ICME and NeurIPS, Senior PC Member of AAAI and IJCAI. 
\end{IEEEbiography}
\vspace{-10mm}

\begin{IEEEbiography}[{\includegraphics[width=1in,height=1.25in]{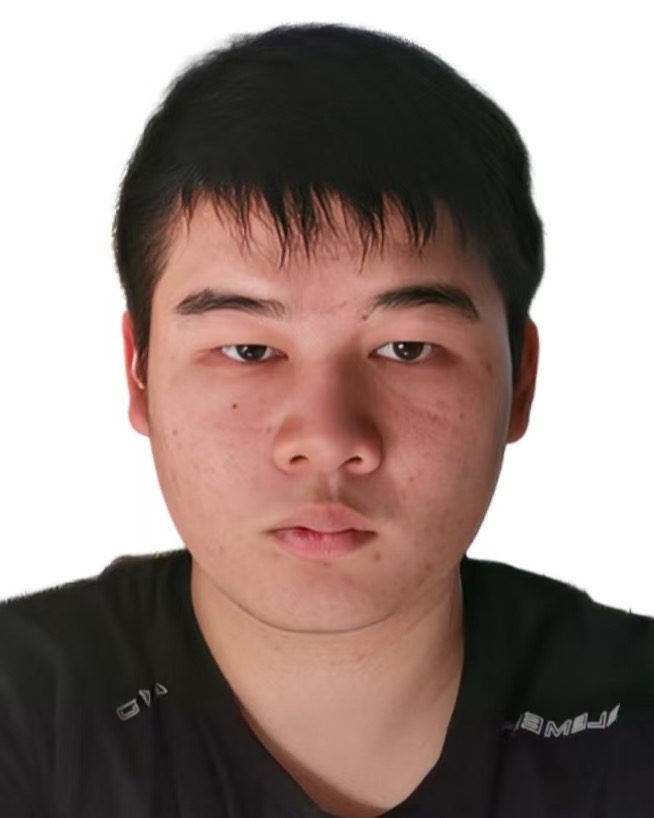}}] {Zijian Fu} received the Bachelor of Engineering degree from the Beijing University of Posts and Telecommunications, China, in 2024, where he is currently pursuing the Master’s degree. His research interests include scene graph generation, multimodal learning, and autonomous driving.
\end{IEEEbiography}
\vspace{-10mm}

\begin{IEEEbiography}[{\includegraphics[width=1in,height=1.25in]{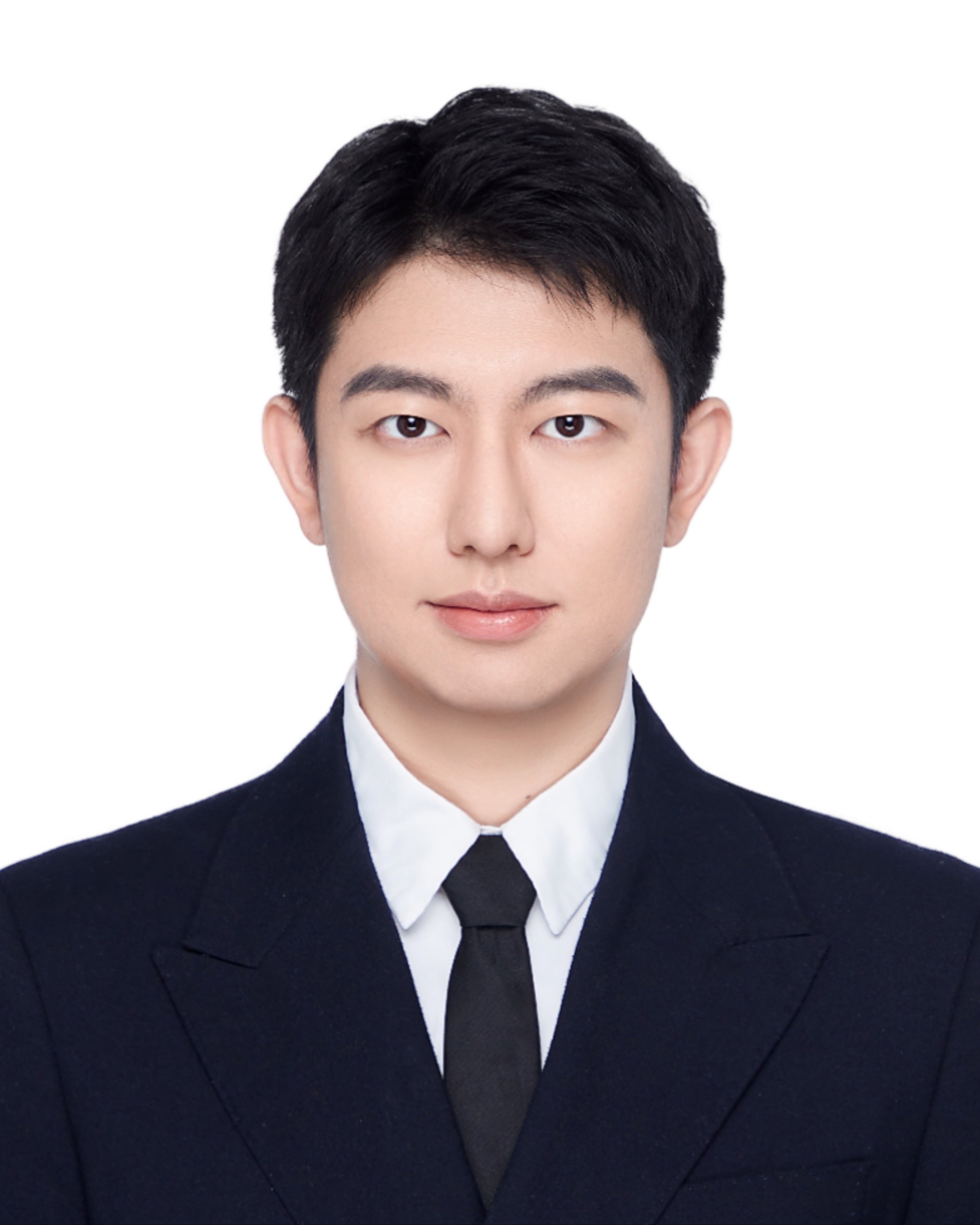}}] {Changsheng Lv} received the Bachelor of Engineering degree from the Beijing University of Posts and Telecommunications, China, in 2021, where he is currently pursuing the Ph.D degree. His research interests include scene graph generation, multimodal learning, and autonomous driving.
\end{IEEEbiography}
\vspace{-10mm}

\begin{IEEEbiography}[{\includegraphics[width=1in,height=1.25in]{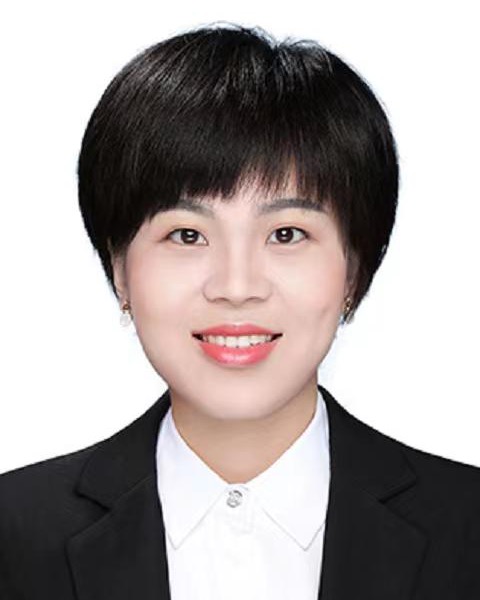}}] {Xianlin Zhang} is now a Associate Professor at Beijing University of Posts and Telecommunications, China. Her current research focuses on generative artificial intelligence and video analysis and understanding. She received her Ph.D. degree from Beijing University of Posts and Telecommunications in 2019. To date, she have authored or co-authored more than 20 publications in prestigious journals and conferences.
\end{IEEEbiography}
\vspace{-10mm}

\begin{IEEEbiography}[{\includegraphics[width=1in,height=1.25in]{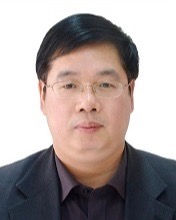}}]{Huadong Ma} (Fellow, IEEE) received the Ph.D. degree in computer science from the Institute of Computing Technology, Chinese Academy of Science, Beijing, China, in 1995. He is currently a Professor of School of Computer Science, Beijing University of Posts and Telecommunications, Beijing, China. His current research interests include Internet of Things, sensor networks, and multimedia computing. He has authored more than 300 papers in ACM/IEEE Transactions or conferences. He was the recipient of the Natural Science Award of the Ministry of Education, China, in 2017, 2019 Prize Paper Award of IEEE TRANSACTIONS ON MULTIMEDIA, 2018 Best Paper Award from IEEE MULTIMEDIA, Best Paper Award in IEEE ICPADS 2010, Best Student Paper Award in IEEE ICME 2016 for his coauthored papers, and National Funds for Distinguished Young Scientists in 2009. He was/is an Editorial Board Member of the IEEE TRANSACTIONS ON MULTIMEDIA, IEEE INTERNET OF THINGS JOURNAL, ACM Transactions on Internet of Things. He is the Chair of ACM China.
\end{IEEEbiography}

\end{document}


\maketitle
\setcounter{page}{1}

This Supplementary Material contains three components: details of the proposed Multi-Modal Scene Graph Dataset in Section~\ref{Sec: Details of Multi-Modal Scene Graph Dataset}, additional implementation details in Section~\ref{Sec: Additional Implementation Details}, and further visualization results in Section~\ref{Sec: Further Visualization Results}.

\section{Details of the Multi-Modal Scene Graph Dataset}
\label{Sec: Details of Multi-Modal Scene Graph Dataset}

This section provides the data preparation and annotation details that
are omitted from the main paper for conciseness. We first describe the
video normalization, temporal segmentation, and ontology, and then
present the three-stage foundation-model annotation pipeline and the
subsequent verification procedure. We also report the annotation cost,
the effect of annotation correction, and the complete dataset statistics.
All definitions and numerical conventions in this appendix follow the
current task formulation in the main paper.

\subsection{Data Preparation Details}
\label{Sec: Data Preparation}

\noindent\textbf{Video Source.}
Our dataset is built upon all $9{,}288$ one-minute musical performance
videos of MUSIC-AVQA~\cite{li2022learning}, which comprise $7{,}422$
real recordings and $1{,}866$ synthesized ensembles and cover solo,
duet, and multi-instrument performances of $22$ instrument categories.
Each video provides temporally aligned visual frames and a monaural
audio track resampled to $16$\,kHz. Videos shorter than $60$ seconds
are extended to the target duration by repeating the final second of
audio, such that every video is normalized to a canonical $60$-second
timeline.

\noindent\textbf{Temporal Segmentation.}
Each normalized video is uniformly divided into $T{=}10$
non-overlapping $6$-second segments, and one multi-modal scene graph is
annotated for each segment, yielding $92{,}880$ annotation units in
total. For segment-level annotation, every segment is represented by a
key frame sampled at its temporal center together with the
corresponding $1$-second audio excerpt. The final refinement stage,
however, operates on the entire video and its synchronized audio stream
in a single pass.

\noindent\textbf{Ontology.}
The annotation ontology instantiates the entity vocabulary
$\mathcal{O}$ and the relation vocabulary $\mathcal{R}$. The entity
vocabulary comprises $25$ categories: the $22$ instrument categories
inherited from MUSIC-AVQA~\cite{li2022learning}, an additional
\textit{other\_instrument} category for instruments beyond this
taxonomy, and the general categories \textit{person} and
\textit{scene}. The latter represents the global visual context of a
segment.

The relation vocabulary consists of six predicates in three
modality-specific groups, with explicit type constraints on admissible
subjects and objects. The spatial relations (\textit{left},
\textit{middle}, and \textit{right}) hold between an instrument and the
scene and describe its horizontal position in the frame. The action
relations (\textit{play} and \textit{hold}) hold between a person and
an instrument. The former requires audible evidence that the
instrument is sounding, whereas the latter denotes physical contact
without acoustic evidence. The auditory relation \textit{louder}
compares the sound intensity of two sounding instruments and is
determined from the audio track. These constraints define three
admissible triplet forms: instrument--spatial--scene,
person--action--instrument, and instrument--auditory--instrument. They
yield $644$ admissible triplet categories. Together with the
\textit{nothing} label for segments without a valid relation, the final
label space contains $645$ categories.

\subsection{Annotation Pipeline Details}
\label{Sec: Data Annotation}

\begin{figure*}[t]
\centering
\includegraphics[width=0.98\textwidth]{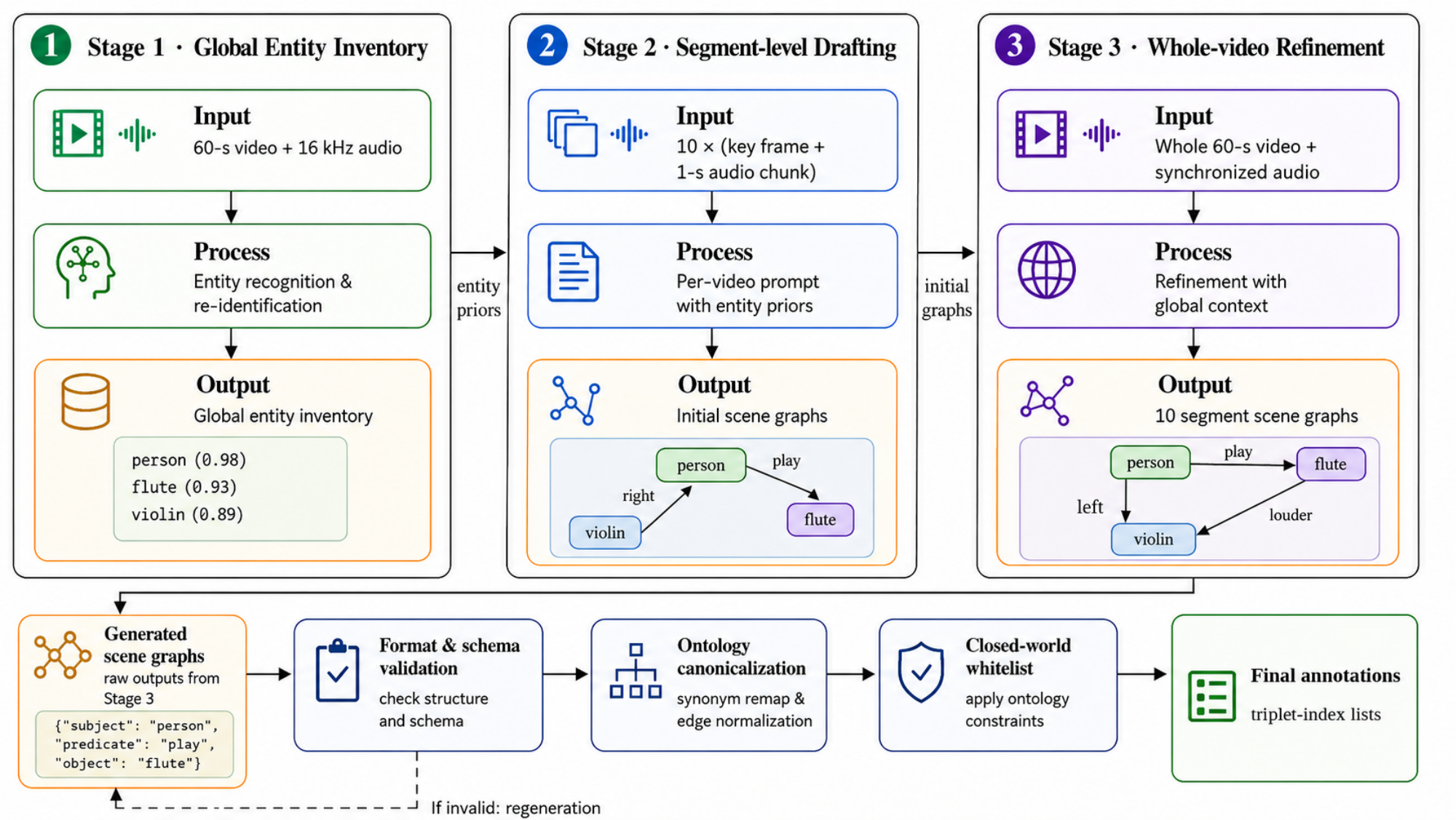}
\caption{Detailed construction pipeline for the multi-modal scene graph
dataset. Stage~1 extracts a global entity inventory with persistent
instance identities from the complete video. Stage~2 conditions
MiniCPM-o~2.6 on this inventory and drafts one scene graph for each
$6$-second segment from its key frame and synchronized audio. Stage~3
uses Qwen3-Omni-30B-A3B (Thinking) to refine all ten segment graphs with
whole-video audio--visual context. The resulting outputs are subjected
to format and schema validation, ontology canonicalization, and a
closed-world whitelist; invalid outputs are regenerated before the
triplet-index annotations are stored.}
\label{Fig: annotation_pipeline_appendix}
\end{figure*}

The annotation procedure contains three cascaded stages performed by
multi-modal foundation models, followed by an automatic
verification-and-regeneration loop, as summarized in
Figure~\ref{Fig: annotation_pipeline_appendix}.

\noindent\textbf{Stage~1: Entity Inventory and Re-Identification.}
A vision--language model processes the complete video and produces a
global entity inventory containing all persons and instruments. Each
entity is assigned a persistent integer identifier, a category from the
ontology, and a confidence score. The prompt imposes instance-level
constraints: each physical object corresponds to one entity, distinct
instances of the same category are not merged, and identities are
preserved across occlusions and shot transitions. It also specifies
visual criteria for disambiguating easily confused instruments, such
as \textit{erhu} and \textit{violin}.

\noindent\textbf{Stage~2: Segment-Level Graph Generation.}
The entity inventory is incorporated into a scene-graph instruction
template to create a customized prompt for each video. Conditioned on
this prompt, MiniCPM-o~2.6~\cite{yao2024minicpm} generates an initial
scene graph for each segment from its key frame and audio excerpt. The
prompt follows an audio-over-video precedence principle: when the two
modalities disagree, the audio evidence determines whether a
\textit{play} event occurs and determines all \textit{louder}
comparisons. It also requires every visible instrument to have exactly
one spatial relation, retains a person only when it participates in a
confirmed action relation, permits at most one \textit{louder}
relation per segment, and removes entities or relations whose
confidence is below $0.5$.

\noindent\textbf{Stage~3: Whole-Video Refinement.}
Qwen3-Omni-30B-A3B (Thinking)~\cite{xu2025qwen3} receives the complete
$60$-second video, its audio stream, and one instruction covering all
ten segments. It then regenerates the complete sequence of segment-level
scene graphs in one inference pass. The model performs explicit
chain-of-thought reasoning, but the intermediate traces are discarded
before the annotations are stored. Whole-video processing helps
maintain consistent entity identities and correct residual errors in
the initial segment-level graphs, which are used only as auxiliary
references.

\noindent\textbf{Verification, Regeneration, and Canonicalization.}
The generated graphs are processed by a correction pipeline combining
rule-based format constraints, LLM-based question-relevance validation,
and expert correction on a sampled subset. Outputs that are truncated,
empty, or fail to cover all ten segments are re-queued for Stage~3
until a well-formed result is obtained. Videos with more than two empty
segments are re-annotated, with the Stage~2 graphs used as a fallback
when necessary.

The validated graphs then undergo rule-based canonicalization.
Out-of-vocabulary or synonymous categories are mapped to the ontology
(for example, \textit{electric\_guitar} is mapped to
\textit{acoustic\_guitar}, and \textit{viola} to \textit{violin});
predicates are normalized; spatial relations mistakenly anchored on a
person are reassigned to the instrument being played; and every
remaining triplet is checked against the admissible forms in
Section~\ref{Sec: Data Preparation}. Violating triplets are discarded,
and segments without an admissible triplet receive the
\textit{nothing} label. Finally, an LLM checks whether the relations are
consistent with the question--answer pairs associated with the video,
and human experts inspect and correct a sampled subset. The
question--answer pairs are used only as post-hoc consistency evidence;
they are never supplied as conditioning inputs for graph generation and
do not change the question-independent definition of the annotation
task. The final annotations are serialized as per-video sequences of
ten segment-level triplet sets.

\subsection{Cost and Quality of Foundation-Model Annotations}
\label{Sec: Annotation Cost and Quality}

The scene graph annotations are generated entirely offline using eight
RTX~4090 GPUs, with an average processing time of approximately three
minutes per video. This cost is incurred only once when constructing
the training annotations. No multi-modal large language model is used
during inference, so the runtime overhead of the final AVQA model
remains comparable to QA-TIGER~\cite{2025tiger}.

The raw foundation-model outputs are not used directly. They are
processed by the rule-based validation, LLM-based relevance checking,
and sampled expert correction described above. As shown in
Table~\ref{tab:ann_correction}, training SHRIKE with the corrected
annotations increases the average accuracy from $61.52\%$ to
$62.42\%$, with the largest improvement on visual questions. This
comparison verifies that the correction pipeline improves the utility
of the generated supervision.

\begin{table}[t]
    \centering
    \footnotesize
    \caption{Ablation study on annotation correction.}
    \label{tab:ann_correction}
    \renewcommand{\arraystretch}{1.2}
    \setlength{\tabcolsep}{3pt}
    \begin{tabular*}{\columnwidth}{@{\extracolsep{\fill}} l|cccc}
        \toprule
        \textbf{Method} & \textbf{A-QA} & \textbf{V-QA} & \textbf{AV-QA} & \textbf{Average} \\
        \midrule
        SHRIKE w/ raw ann.  & 66.00 & 59.34 & 61.38 & 61.52 \\
        SHRIKE w/ corr ann. & 66.00 & 61.54 & 61.90 & \textbf{62.42} \\
        \bottomrule
    \end{tabular*}
\end{table}

\subsection{Dataset Statistics}
\label{Sec: Dataset Statistics SM}

Table~\ref{Table: dataset statistics SM} reports the complete dataset
statistics that complement the distribution plots and temporal example
in the main paper. The dataset contains $9{,}288$ videos and $92{,}880$
annotated $6$-second segments, with $461{,}292$ relation triplets in
total and an average of $4.97$ triplets per segment. Spatial, action,
and auditory relations contribute $193{,}754$, $188{,}884$, and
$76{,}764$ instances, respectively. Within the action relations,
\textit{play} is substantially more frequent than \textit{hold}
($188{,}324$ versus $560$), reflecting the performance-oriented
filtering used during segment-level drafting.

Among the $644$ admissible triplet categories, $560$ are observed in
the final annotations. Only $1{,}896$ segments ($2.0\%$) receive the
\textit{nothing} label, while the auditory relation \textit{louder}
appears in $6{,}718$ videos ($72.3\%$). These results indicate that the
dataset provides dense structural supervision and broad coverage of
spatial, action, and cross-modal auditory relations.

\begin{table}[t]
\centering
\caption{Detailed statistics of the multi-modal scene graph dataset.}
\label{Table: dataset statistics SM}
\setlength{\tabcolsep}{1.2mm}
\begin{tabular}{lr}
\toprule
\textbf{Item} & \textbf{Value} \\
\midrule
Videos (real / synthetic) & $9{,}288$ ($7{,}422$ / $1{,}866$) \\
Video length / segments per video & $60$\,s / $10 \times 6$\,s \\
Annotated segments & $92{,}880$ \\
Entity categories (instruments) & $25$ ($23$) \\
Predicates (spatial / action / auditory) & $6$ ($3/2/1$) \\
Triplet label space & $645$ (incl.\ \textit{nothing}) \\
Triplet instances & $461{,}292$ \\
\quad spatial / action / auditory & $193{,}754$ / $188{,}884$ / $76{,}764$ \\
Avg.\ triplets per segment & $4.97$ \\
Observed triplet categories & $560$ \\
Videos with \textit{louder} relations & $6{,}718$ ($72.3\%$) \\
Empty (\textit{nothing}) segments & $1{,}896$ ($2.0\%$) \\
\bottomrule
\end{tabular}
\end{table}

\section{Additional Implementation Details}
\label{Sec: Additional Implementation Details}
In this section, we present additional implementation details, including the patch merging method for visual tokens (Section~\ref{Sec: Patch Merge}), specific aspects of the model's training strategy (Section~\ref{Sec: Training Strategy}),Comparison of inference overhead(Section~\ref{Sec: Comparison of inference overhead}) Runtime and Parameter Efficiency of SHRIKE(Section~\ref{Sec: Runtime and Parameter Efficiency}) and implementation details of the KAN experts (Section~\ref{Sec: Details of the KAN Experts}).
\subsection{Patch Merge}
\label{Sec: Patch Merge}
We adopt the patch merging approach introduced in ToMe~\cite{2022tome}, where visually similar tokens are progressively combined at each transformer layer. This process partitions tokens into groups, computes pairwise similarities, and then applies an averaging mechanism to produce fused token representations. Incorporating this strategy ensures that our feature extraction is consistent with baseline~(TSPM~\cite{li2024boosting}), maintaining alignment across the experimental design.
\subsection{Training Strategy}
\label{Sec: Training Strategy}
In our model design, we argue that the prediction of scene graphs by the Multi-Modal Scene Graph Decoder (MMSG) should remain independent of the specific question being asked. To this end, we adopt a two-stage training strategy. In the first stage, we pre-train the MMSG module separately. Specifically, the input to the module consists of visual and audio features, while the output is a set of classification probabilities over triplets representing the scene graph for each video frame.

To optimize the model, we employ the Hungarian loss, which effectively handles permutation-invariant triplet matching. During training, we assume a fixed number of 12 triplets per clip. If the number of valid triplets is fewer than 12, we pad the missing entries with a special \textit{None} token.  Through this procedure, we obtain a robust MMSG module capable of accurately predicting multi-modal scene graph structures without being influenced by question-specific information. To prevent the model from overfitting to padded entries, we assign a lower loss weight of 0.3 to None triplets, while valid triplets are assigned a weight of 1. The model is trained for 15 epochs using the Adam optimizer with an initial learning rate of 1$e^{-4}$, which is decayed by a factor of 0.1 every 5 epochs.

In the second stage, we freeze the parameters of the pre-trained MMSG module and train the remaining components of the model. At this stage, the model takes visual and audio features, along with the question embedding, as input, and predicts the answer by outputting classification probabilities over candidate answers. We use the cross-entropy loss to supervise the training. The model is trained for 15 epochs using the Adam optimizer with an initial learning rate of 
1$e^{-4}$, which is decayed by a factor of 0.1 every 9 epochs to stabilize training and facilitate convergence.

We also evaluated all models under different encoder backbones. As shown in Fig~\ref{dif encoder}, our method consistently achieves the best performance across all encoder configurations
\begin{table}[t]
\centering
\caption{Performance with different encoders serving both visual and text feature extraction.}
\label{dif encoder}
\setlength{\tabcolsep}{6pt}
\renewcommand{\arraystretch}{1.12}
\resizebox{\columnwidth}{!}{%
\begin{tabular}{l c c c c c}
\toprule
Method & CLIP Encoder & A-QA & V-QA & AV-QA & Avg \\
\midrule
PSTP-Net\cite{li2023progressive} & B/32 & 70.91 & 77.26 & 72.57 & 73.52 \\
TSPM\cite{li2024boosting}     & B/32 & 76.91 & 81.92 & 72.57 & 75.81 \\
QA-TIGER\cite{2025tiger}     & B/32 & 76.23 & \textbf{84.10} & 72.14 & 76.03 \\
\textbf{Ours} & B/32 & \textbf{78.15} & 83.86 & \textbf{72.23} & \textbf{76.36} \\
\bottomrule
\end{tabular}
}
\end{table}

\begin{algorithm}
\caption{Gaussian KAN Experts Module}\label{algo:KAN_experts}
\begin{algorithmic}
    \State \textbf{Input:} Question features: $q_s \in \mathbb{R}^{D}$. Video features: $v_q \in \mathbb{R}^{T \times D}$. Audio features: $a_q \in \mathbb{R}^{T \times D}$. Patch features : $p_v,p_a \in \mathbb{R}^{T \times D}$. 
    \State \textbf{Output:} Temporal integrated features:  Aggregated visual-related patch features $\widetilde{F_{p_v}} \in \mathbb{R}^{D}$, aggregated audio-related patch features $\widetilde{F_{p_a}} \in \mathbb{R}^{D}$. Aggregated audio features $\widetilde{F_{a}} \in \mathbb{R}^{D}$.

    \vspace{0.5em}
    \State \textbf{Initialization:} Initialize the center of $E$ experts to the central positions of $E$ segments.
    \State\quad $\text{margin} = \dfrac{1}{2E}$ 
    \begin{algorithmic}
    \For{$i = 0$ \textbf{to} $E - 1$}
        \State $\mu_{\text{fixed}}[i] = \text{margin} + i \cdot \frac{1 - 2 \cdot \text{margin}}{E - 1}$
    \EndFor
    \end{algorithmic}
    
    \vspace{0.5em}
    \State \textbf{ Question-driven  Attention:}
    \State \quad $v^{\prime}_q \: = \text{CA}(q_s, v_q, v_q),  \: a^{\prime}_q \:=\text{CA}(q_s, a_q, a_q)$
    
    \vspace{0.5em}
    \State \textbf{Experts Router:}
\State \quad $\begin{aligned}[t]
r_v &= \operatorname{Softmax}(\operatorname{Linear}(\mathbf v'_q)),\\
r_a &= \operatorname{Softmax}(\operatorname{Linear}(\mathbf a'_q))
\end{aligned}$

    \vspace{0.5em}
    \State \textbf{ Gaussian Weight Generation:}
    \State \quad $\mu_{\text{offset}|v}, \sigma_{v} \:= \text{FC}(v^{\prime}_q),\:\:\mu_{\text{offset}|a}, \sigma_{a} \:= \text{FC}(a^{\prime}_q)$
    \State \quad Adjust centers and normalize widths:
    \State \quad \textbf{for} $i = 0$ to $E-1$ \textbf{do}
    \State \quad  \quad $\mu_{v}[i] \:= \mu_{\text{fixed}}[i] + \text{Tanh}(\mu_{\text{offset}|v}[i]) \cdot \text{margin}$
    \State \quad \quad $\mu_{a}[i] \:= \mu_{\text{fixed}}[i] + \text{Tanh}(\mu_{\text{offset}|a}[i]) \cdot \text{margin}$
    \State \quad\quad $\begin{aligned}[t]
\sigma_v[i] &= \operatorname{sigmoid}(\sigma_v[i]),\\
\sigma_a[i] &= \operatorname{sigmoid}(\sigma_a[i])
\end{aligned}$
    \State \quad \textbf{end for}
    \State \quad Generate temporal Gaussian weights:
    \State \quad \textbf{for} $i = 0$ to $E-1$ \textbf{do}
    \State \quad\quad $\begin{aligned}[t]
\delta_v[i] &= \mathcal{N}(\mu_v[i], \sigma_v[i]^2),\\
\delta_a[i] &= \mathcal{N}(\mu_a[i], \sigma_a[i]^2),
\end{aligned}$
    \State \quad \quad $\delta_{v}[i] \:  =\frac{\delta_{v}[i]}{\max(\delta_{v}[i])},\:\delta_{a}[i] \:= \frac{\delta_{a}[i]}{\max(\delta_{a}[i])}$
    \State \quad \textbf{end for}

    \vspace{0.5em}
    \State \textbf{ Integration of KAN Experts Output:}
    \State \quad  $p_v = \mathbf{v}_q + p_v$
    \State \quad  $p_a = \mathbf{v}_q + p_a$
    \vspace{0.2em}
    \State \quad  $ \widetilde{F_{a}} = \sum_{i=0}^{E-1} \delta_{a}[i]\cdot r_{a}[i]\cdot \mathcal{F}_{\text{KAN}}^{i}(a_q)$
    \State \quad$\widetilde{F_{p_v}} = \sum_{i=0}^{E-1} \delta_v[i]\, r_v[i]\, \mathcal{F}_{\text{KAN}}^{i}(p_v)$
\State \quad$\widetilde{F_{p_a}} = \sum_{i=0}^{E-1} \delta_v[i]\, r_v[i]\, \mathcal{F}_{\text{KAN}}^{i}(p_a)$

    \vspace{0.2em}
    \State \quad  \Return $\widetilde{F_{a}},\:\widetilde{F_{p_v}},\:\widetilde{F_{p_a}}$
    
\end{algorithmic}
\end{algorithm}

\subsection{Comparison of inference overhead}
\label{Sec: Comparison of inference overhead}
For inference, SHRIKE runs 33s/batch (bs=32) on an RTX 3090 while QA-TIGER runs 30s. In return, SHRIKE yields clear gains on challenging audio-spatial reasoning, \emph{e.g.}, ``Where is the first sounding instrument?'' 42.44\% vs. 33.72\% for QA-TIGER showing the good trade-off.
\subsection{Runtime and Parameter Efficiency}
\label{Sec: Runtime and Parameter Efficiency}
On the Music-AVQA dataset,our SHRIKE model requires about 3 minutes per epoch during the first training stage (10 epochs total), and about 6 minutes per epoch during the second stage (15 epochs total), resulting in a total training time of roughly 2 hours. For inference, SHRIKE takes around 33 seconds on an NVIDIA RTX 3090 GPU (batch size 32).

\subsection{Details of the KAN Experts}
\label{Sec: Details of the KAN Experts}
Inspired by QA-TIGER~\cite{2025tiger}, we initialize the Gaussian centers of experts $\mu_{fixed}$ as $E$ points spaced at fixed intervals as shown in algorithm~\ref{algo:KAN_experts}. By applying learnable offsets to the fixed positions, the centers are adjusted within restricted margins. This guarantees that each expert attends to unique temporal intervals, improving the effectiveness of capturing relevant segments and minimizing redundancy, thereby maintaining the specialization of each expert.


\section{Further Visualization Results}
\label{Sec: Further Visualization Results}
\begin{figure*}[t]
    \centering
    \begin{minipage}[b]{0.9\textwidth}
        \centering
        \includegraphics[width=1.0\textwidth]{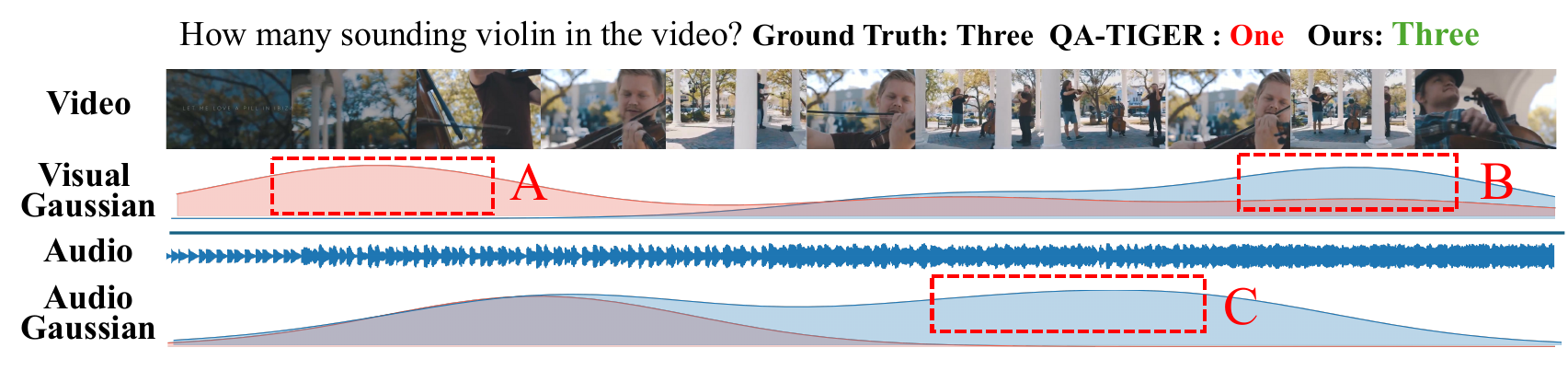}
        \vspace{-6mm}
        \subcaption{Question 1}
        \label{qualitative4}
    \end{minipage}
    \par\vspace{1.5mm}
    \begin{minipage}[b]{0.9\textwidth}
        \centering
        \includegraphics[width=1.0\textwidth]{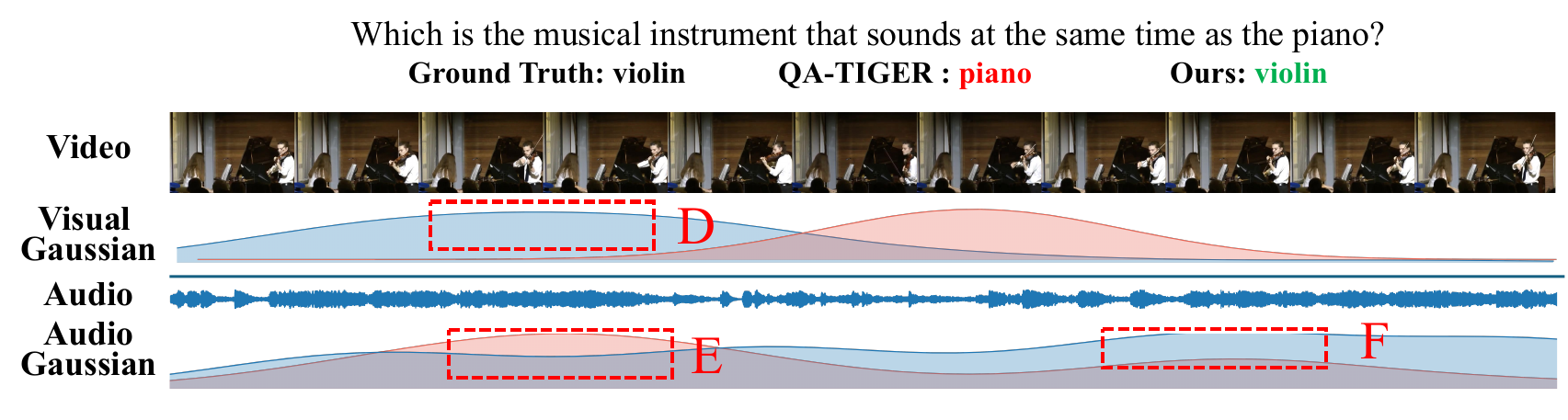}
        \vspace{-6mm}
        \subcaption{Question 2}
        \label{qualitative5}
        
    \end{minipage}
    \par\vspace{1.5mm}
    \begin{minipage}[b]{0.9\textwidth}
        \centering
        \includegraphics[width=1.0\textwidth]{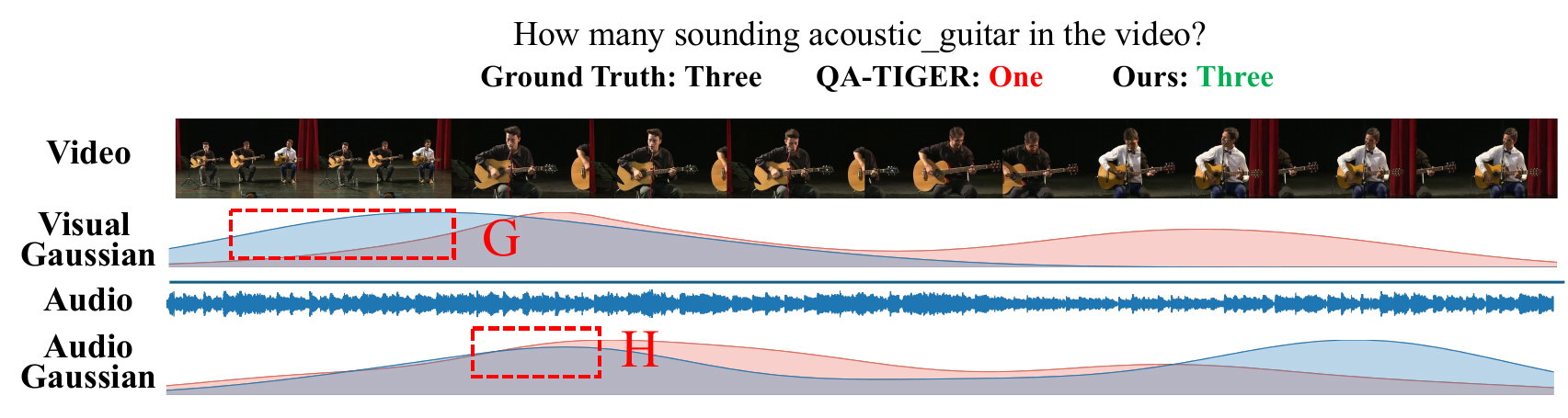}
        \vspace{-6mm}
        \subcaption{Question 3}
        \label{qualitative6}
    \end{minipage}
    \par\vspace{1mm}
    \caption{Comparison of key temporal segment capture between SHRIKE and QA-TIGER.
    \textcolor{red}{Red} wavy lines represent QA-TIGER, and \textcolor{blue}{blue} wavy lines represent SHRIKE.}
    \label{Fig:visuailzation_sup}
    \vspace{-2mm}
\end{figure*}

\begin{figure*}[t]
    \centering
    \includegraphics[width=0.85\linewidth]{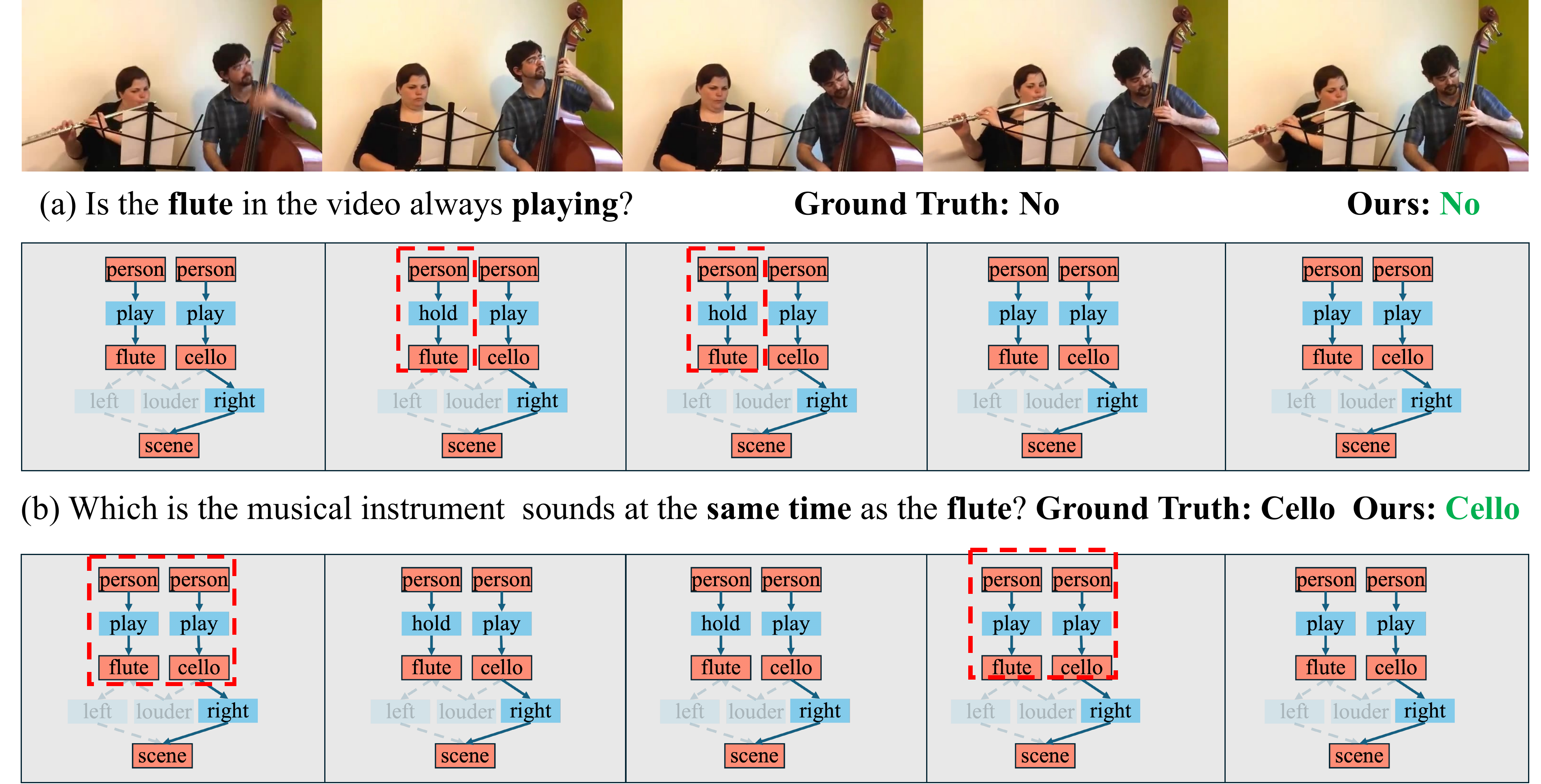}
    \vspace{-1mm}
    \caption{Visualized relationship triplet selection result. In the given example, we convert all the relationship triplets into a scene graph and highlight the triplets selected by our method.}
    \label{Fig:REL2}
    \vspace{-3mm}
\end{figure*}

As shown in Figure~\ref{Fig:visuailzation_sup} and Figure~\ref{Fig:REL2}, we present more visualization results comparing our proposed
method with other baseline models. It can be seen from the figures that our proposed method
outperforms the original method.

In Figure~\ref{qualitative4}, SHRIKE's visual Gaussian attention correctly focuses on the frame where three violins appear simultaneously in Region B, whereas QA-TIGER misplaces its attention on an incorrect frame in Region A. In addition to visual attention, SHRIKE's audio Gaussian mechanism accurately focuses on the segment where three violins are playing at the same time, highlighting its superior multi-modal reasoning ability in Region C. In Figure~\ref{qualitative5}, SHRIKE's audio Gaussian attention broadly attends to all audio frames, which aligns with the fact that the piano and violin are playing simultaneously throughout the entire clip in Region E and Region F. For Figure~\ref{qualitative6}, SHRIKE demonstrates a more accurate visual understanding by first identifying the frame in which three acoustic guitars are present simultaneously. QA-TIGER, on the other hand, mistakenly attends to a frame where only a person is present in the scene graph, which ultimately results in an incorrect prediction.

\begin{figure*}[t]
    \centering
    \begin{minipage}[b]{0.9\textwidth}
        \centering
        \includegraphics[width=1\textwidth]{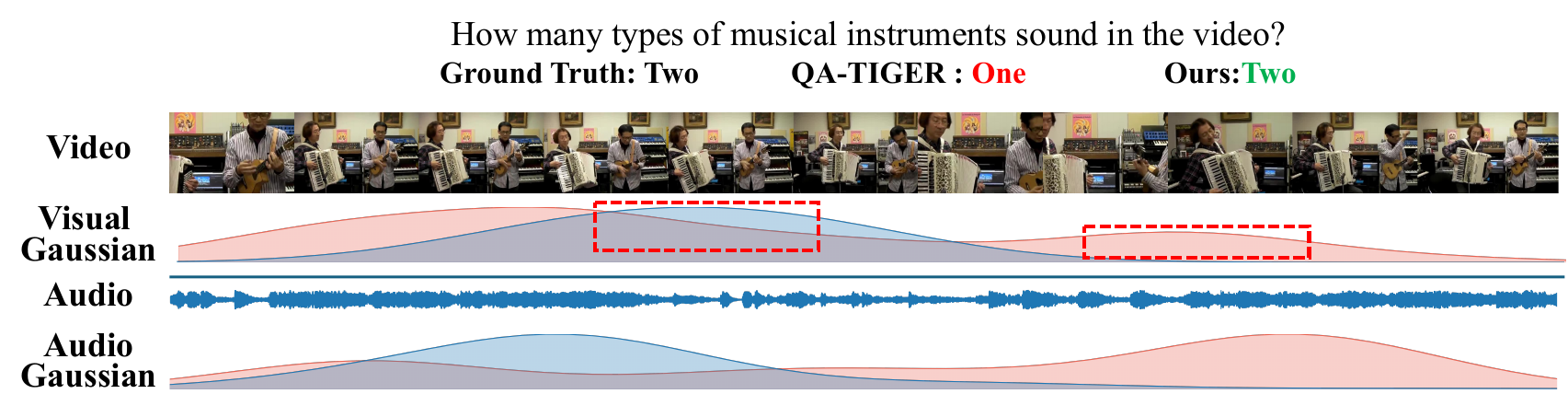}
        \vspace{-6mm}
        \subcaption{Question 1}
        \label{qualitative7}
    \end{minipage}
    \par\vspace{1.5mm}
    \begin{minipage}[b]{0.9\textwidth}
        \centering
        \includegraphics[width=1\textwidth]{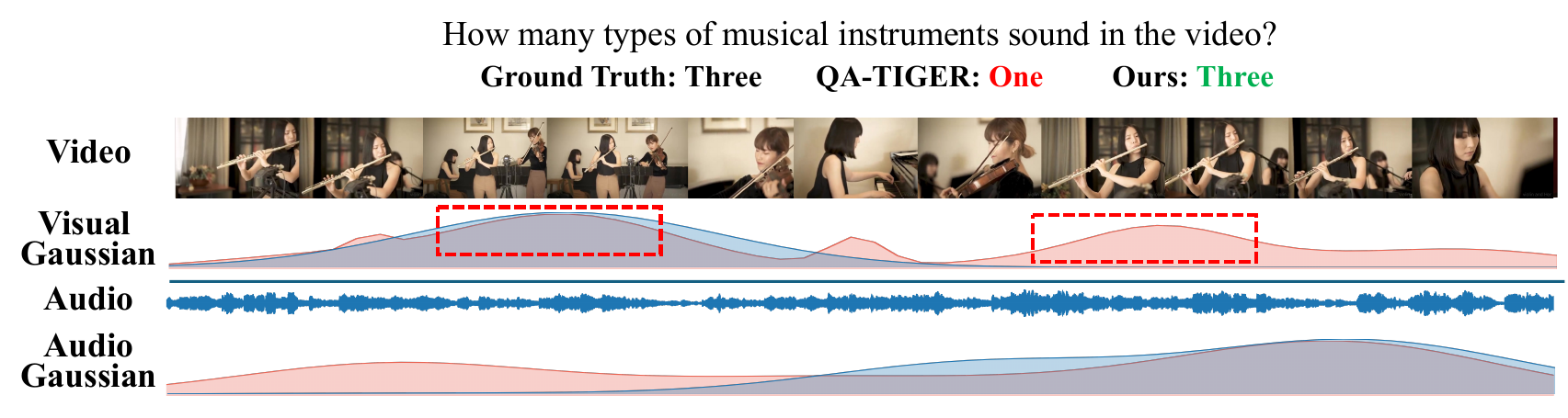}
        \vspace{-6mm}
        \subcaption{Question 2}
        \label{qualitative8}
        
    \end{minipage}
    \par\vspace{1.5mm}
    \begin{minipage}[b]{0.9\textwidth}
        \centering
        \includegraphics[width=\textwidth]{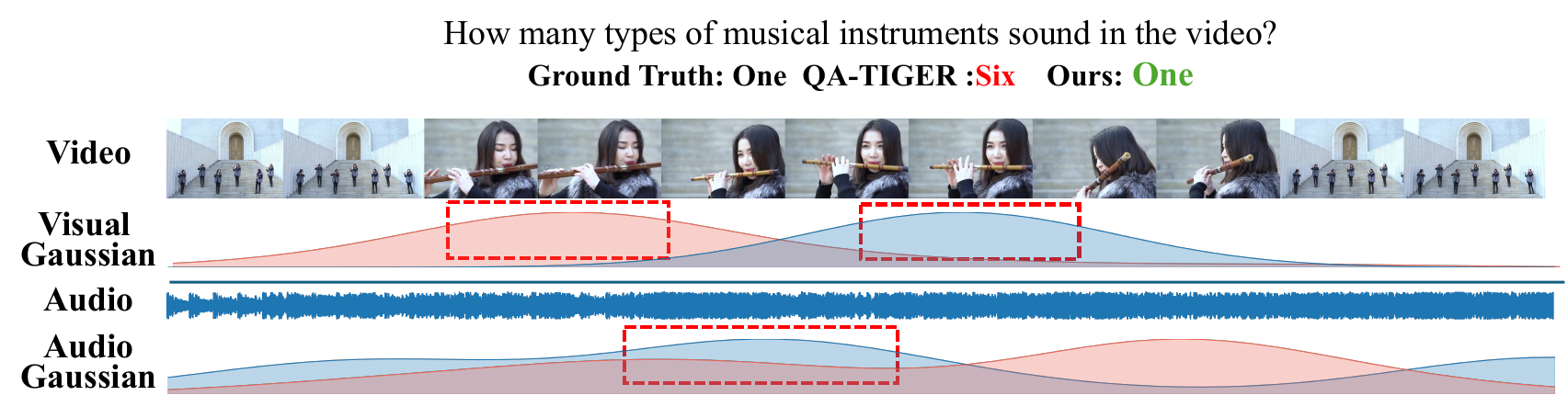}
        \vspace{-6mm}
        \subcaption{Question 3}
        \label{qualitative9}
    \end{minipage}
    \par\vspace{1mm}
    \caption{Comparison of key temporal segment capture between SHRIKE and QA-TIGER.
    \textcolor{red}{Red} wavy lines represent QA-TIGER, and \textcolor{blue}{blue} wavy lines represent SHRIKE.}
    \label{Fig:visuailzation_sup2}
    \vspace{-2mm}
\end{figure*}

\begin{figure*}[t]
    \centering
    \includegraphics[width=0.83\linewidth]{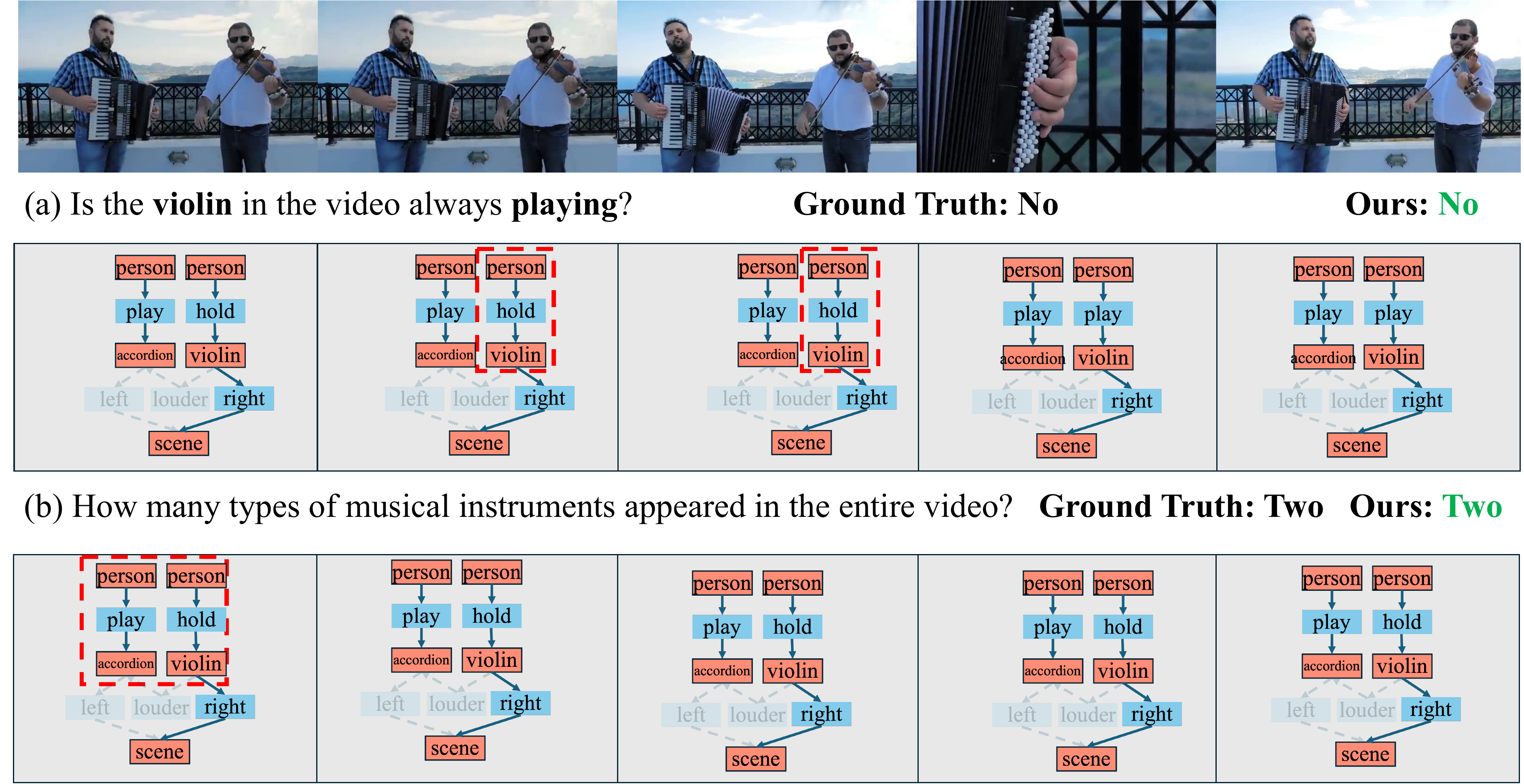}
    \vspace{-1mm}
    \caption{Visualized relationship triplet selection result. In the given example, we convert all the relationship triplets into a scene graph and highlight the triplets selected by our method.}
    \label{Fig:REL3}
    \vspace{-3mm}
\end{figure*}

\begin{figure*}[t]
    \centering
    \begin{minipage}[b]{0.9\textwidth}
        \centering
        \includegraphics[width=1\textwidth]{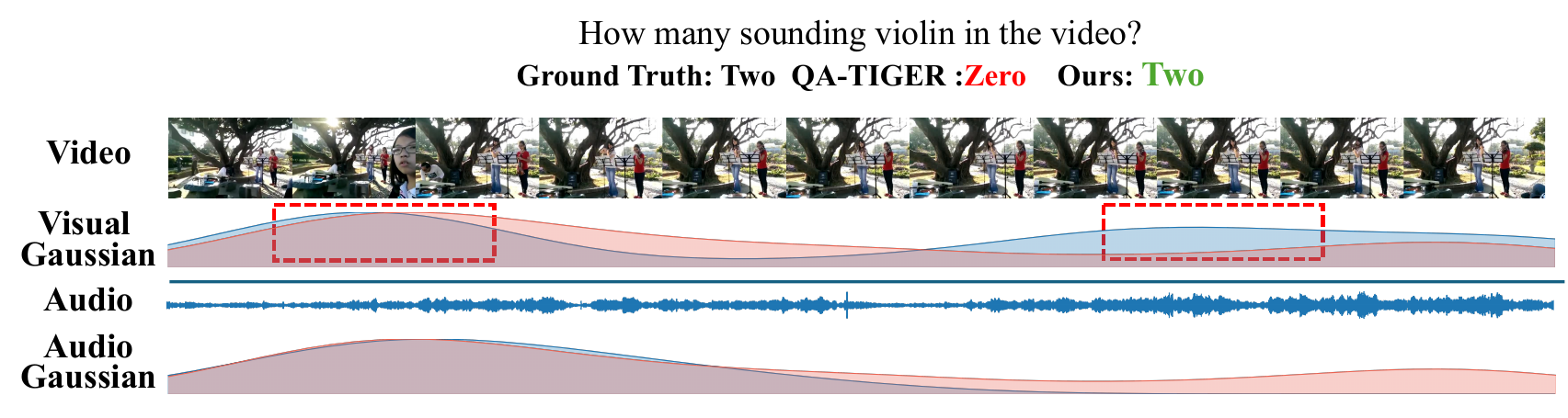}
        \vspace{-6mm}
        \subcaption{Question 1}
        \label{qualitative12}
    \end{minipage}
    \par\vspace{1.5mm}
    \begin{minipage}[b]{0.9\textwidth}
        \centering
        \includegraphics[width=1\textwidth]{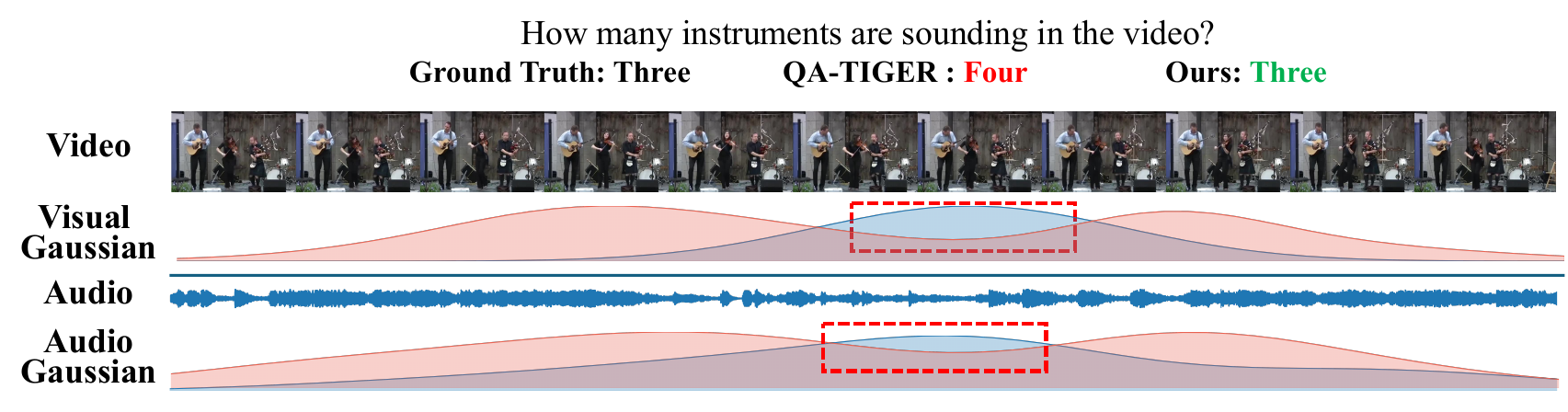}
        \vspace{-6mm}
        \subcaption{Question 2}
        \label{qualitative11}
        
    \end{minipage}
    \par\vspace{1.5mm}
    \begin{minipage}[b]{0.9\textwidth}
        \centering
        \includegraphics[width=1\textwidth]{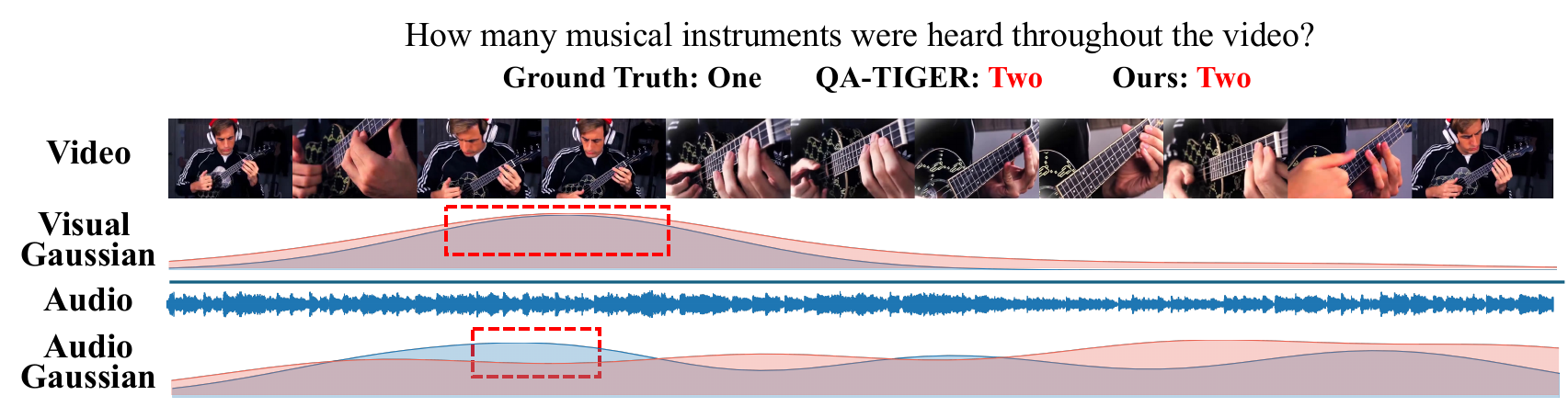}
        \vspace{-6mm}
        \subcaption{Question 3}
        \label{qualitative10}
    \end{minipage}
    \par\vspace{1mm}
    \caption{Comparison of key temporal segment capture between SHRIKE and QA-TIGER.
    \textcolor{red}{Red} wavy lines represent QA-TIGER, and \textcolor{blue}{blue} wavy lines represent SHRIKE.}
    \label{Fig:visuailzation_sup3}
    \vspace{-2mm}
\end{figure*}

\begin{figure*}[t]
    \centering
    \includegraphics[width=0.85\linewidth]{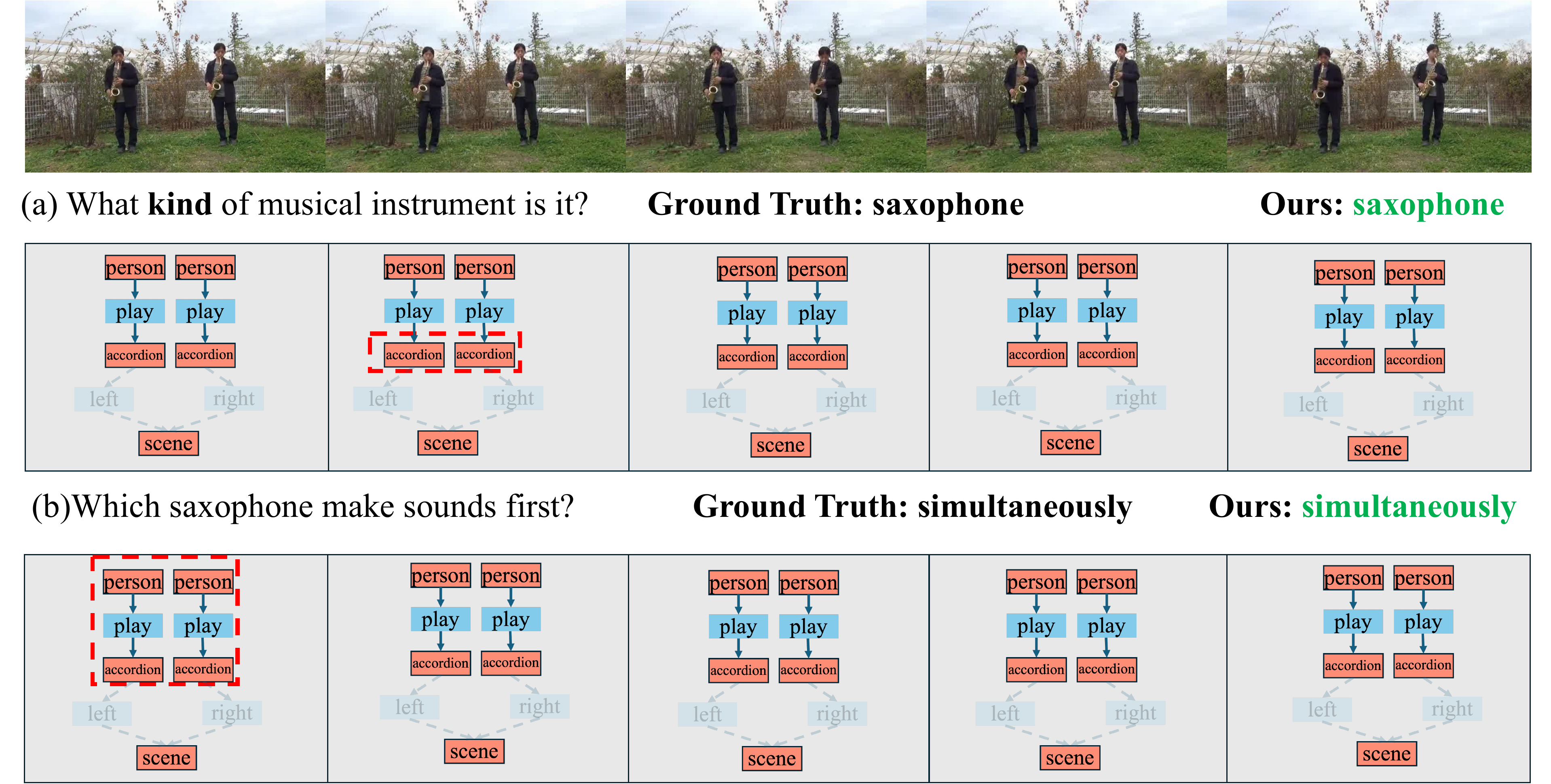}
    \vspace{-1mm}
    \caption{Visualized relationship triplet selection result. In the given example, we convert all the relationship triplets into a scene graph and highlight the triplets selected by our method.}
    \label{Fig:REL4}
    \vspace{-3mm}
\end{figure*}

In Figure~\ref{Fig:REL2}, SHRIKE answers the question by assessing whether the flute is being played at each moment. The model accurately predicts that in the middle clip of the video, the person is not playing the flute. As shown in the figure~\ref{Fig:REL2}(a), it predicts the relation \textit{<person-hold-flute} instead of \textit{<person-play-flute>}. On the other hand, the model accurately predicts both \textit{<person-play-flute>} and \textit{<person-play-cello>} in the video frames, which directly supports answering the question: \textit{"Which musical instrument sounds at the same time as the flute?"} as show in Figure~\ref{Fig:REL2}(b).
We also present more visualization results in Figure~\ref{Fig:visuailzation_sup2}, Figure~\ref{Fig:REL3}, Figure~\ref{Fig:visuailzation_sup3} and Figure~\ref{Fig:REL4}.



\captionsetup[sub]{skip=2pt}

\captionsetup[sub]{skip=2pt}

\captionsetup[sub]{skip=2pt}

\clearpage
\bibliographystyle{IEEEtran}
\bibliography{main}